\documentclass[12pt, fullpage, a4paper]{article}
\usepackage{graphicx, latexsym}
\usepackage{setspace}
\usepackage{amssymb, amsmath, amsthm}
\usepackage{bm}
\usepackage{dsfont}
\usepackage{epstopdf}
\usepackage{rotating} 
\usepackage{booktabs} 
\usepackage{multirow} 
\usepackage{pifont}
\usepackage{floatrow}
\usepackage[natbibapa]{apacite}
\usepackage[svgnames]{xcolor}
\usepackage{hyperref}
\usepackage{cleveref}
\usepackage{statmath}
\usepackage{caption}
\usepackage{subcaption}
\usepackage{authblk}
\usepackage[linesnumbered,ruled,vlined]{algorithm2e}

\hypersetup{
     colorlinks   = true,
     linkcolor    = blue,
     citecolor    = blue
}

\SetCommentSty{mycommfont}
\SetKwInput{KwInput}{Input}                
\SetKwInput{KwOutput}{Output}              
\title{A Semi-supervised CART Model for Covariate Shift}
\author[1]{Mingyang Cai}
\author[1]{Thomas Klausch}
\author[1]{Mark A. van de Wiel}
\date{}
\affil[1]{\small Department of Epidemiology and Data Science, Amsterdam Public Health Research Institute, Amsterdam University Medical Centers Location AMC, Meibergdreef 9, the Netherlands}

\begin{document}

\maketitle

\begin{abstract}
    Machine learning models used in medical applications often face challenges due to the covariate shift, which occurs when there are discrepancies between the distributions of training and target data. This can lead to decreased predictive accuracy, especially with unknown outcomes in the target data. This paper introduces a semi-supervised classification and regression tree (CART) that uses importance weighting to address these distribution discrepancies. Our method improves the predictive performance of the CART model by assigning greater weights to training samples that more accurately represent the target distribution, especially in cases of covariate shift without target outcomes. In addition to CART, we extend this weighted approach to generalized linear model trees and tree ensembles, creating a versatile framework for managing the covariate shift in complex datasets. Through simulation studies and applications to real-world medical data, we demonstrate significant improvements in predictive accuracy. These findings suggest that our weighted approach can enhance reliability in medical applications and other fields where the covariate shift poses challenges to model performance across various data distributions.
\end{abstract}
\section{Introduction}
\label{sec1}
Machine learning models based on trees, such as such as classification and regression trees (CART) \citep{breiman2017classification}, random forests \citep{breiman2001random}, gradient boosting machines \citep{ke2017lightgbm, chen2016xgboost}, and Bayesian additive regression trees \citep{chipman2010bart}, are popular due to their good predictive performance and versatility across a range of tasks. These models partition the feature space into rectangular cells, effectively capturing complex, non-linear relationships. They can handle various data types without extensive preprocessing, including numerical, categorical, and ordinal variables. As a result, these models are widely used for classification, regression, feature selection, and anomaly detection, making them essential in modern machine-learning methodologies.

When training a tree-based model, the training and test datasets are expected to come from the same distribution and have similar joint probability distributions. However, tree-based models may not generalize well when these distributions have significant differences. The distribution discrepancy can lead to inaccurate predictions, as the models may focus on noise or specific patterns in the training data rather than genuine relationships, resulting in reduced performance on new data. 

Domain adaptation is a transfer learning technique \citep{pan2009survey} that mitigates the side effect of data dependency in machine learning models \citep{farahani2021brief, kouw2019review, weiss2016survey, zhuang2020comprehensive}. The training and test sets are referred to as the source and target domains (or sets) in the domain adaptation setting. It involves using knowledge gained from one dataset (the source set) and applying it to a different dataset (the target set). Domain adaptation allows models to better adapt to new datasets. This approach improves generalization, enhances the performance of the model in new target sets, and reduces the sensitivity to the distribution of the source set, especially in situations with limited source data. Domain adaptation techniques can be a powerful tool for practitioners to improve the performance of their models. Various domain adaptation techniques have been developed, including unsupervised and supervised approaches. We focus on the former, and refer to \citep{duan2009domain, zhuang2011exploiting, tommasi2009more} for examples of supervised domain adaptation, and to \citep{sun2022transboost, segev2016learn} for tree-specific \emph{supervised} domain adaptation algorithms.

In this paper we present a domain adaptation framework for tree-based models. Our approach is innovative in improving the predictive performance of tree-based models when there is a difference in the distribution of covariates between the source and target domains. The proposed method is particularly relevant when the target set's outcomes are not observed. We apply an importance weighting approach, discussed in section \ref{sec3}, to address distribution discrepancy within tree-based models. Specifically, source samples that are more representative of the target set are assigned larger weights during model training, thereby improving overall prediction accuracy.

The issue of model generalization is particularly relevant for longitudinal follow-up when the response variable is not observed in the target data. For example, collecting survival outcome data for cancer patients (e.g., death or deterioration) who receive new treatments in clinical analysis may take a long time. Accurately predicting survival is challenging when the characteristics of patients in real-world settings (i.e., the target) differ significantly from those used to train tree-based prediction models in experimental studies (i.e., the source).This mismatch in patient characteristics between the training and real-world datasets significantly undermines the model's ability to predict survival accurately for new patients in the target. It is essential to understand and mitigate the risks of these challenges.

In the unsupervised domain adaptation setting, the primary objective is to train a model in the source domain that generalizes well to the target domain. Feature-level adaptation techniques play a crucial role in unsupervised domain adaptation \citep{gong2012geodesic, mourragui2023computational, pan2010domain, long2013transfer}. These techniques, known as domain-invariant feature learning and distribution matching, aim to align source data with target data by learning a transformation that extracts consistent feature representations across different domains. These techniques usually involve transforming the original features into a new feature space and then minimizing the differences between domains within this new representation space through an optimization process while preserving the underlying structure of the original data. Such feature-based adaptation is limited to classification. Moreover, it is challenging to interpret the model after transferring the feature space with dimension reduction techniques, as the feature space used for model building is different from the original. Deep neural networks are also widely employed in domain adaption or machine learning tasks such as image classification, sentiment analysis and object recognition \citep{long2015learning, ganin2015unsupervised, hoffman2018cycada}. However, to our knowledge, unsupervised transfer learning has yet to be studied for tree-based models. 

We focus on CART to investigate how importance weights affect the selection of optimal splits, thereby improving prediction accuracy in the target domain. Using CART provides a clear understanding, visualization, and interpretation of how the proposed improvements impact prediction performance. This transparency ensures that any improvements are well-understood and validated, making them more likely to also enhance the performance of ensemble trees, such as random forests. For exploratory purposes, we are also considering bagged trees. Additionally, as demonstrated later, the importance weighting may be applied to other tree-based models, such as generalized linear model tree \citep{zeileis2008model}.

The remainder of our article is organized as follows: In section \ref{sec2}, we provide an overview of CART and illustrate how variations in the distribution of predictor variables may impact its performance. Section \ref{sec3} focuses on the importance weighting method and how to integrate this method with CART and other tree-based models. In section \ref{sec4}, we present simulations; in section \ref{sec5}, we offer an example of our method applied to lymphoma patients. Finally, we conclude with a summary and a discussion in section \ref{sec6}.

\section{Classification and regression trees}
\label{sec2}
\subsection{Classification and regression trees}
Classification and Regression Trees (CART) \citep{breiman2017classification} are an essential tool in machine learning and statistical modelling. CART offers interpretability, visual clarity, and the ability to capture complicated relationships between features and response variables. CART is a recursive algorithm that partitions the data space into homogeneous cells using binary splits. Within each cell, a mean prediction model is fitted. The representation of CART used in this paper is based on the work by \citet{scornet2015consistency}. Let $\Theta$ denotes the generic parameter used for splitting, and let $\mathcal{F}(\Theta)$ be the set of all possible functions $f: \mathbb{R}^p \rightarrow \mathbb{R}$, piecewise constant in each cell. The aim is usually to construct a tree estimate $m_n(X, \Theta):\mathbb{R}^p \rightarrow \mathbb{R}$ minimizing the mean squared error in
\begin{equation}
m_n(X, \Theta) := \underset{f\in \mathcal{F}(\Theta)}{\argmin}\frac{1}{n}\sum_{i = 1}^{n}|f(X_i) - Y_i|^2   
\label{eq1}
\end{equation}
in the training set $\mathcal{D}_s = (X_1, Y_1), \dots, (X_n, Y_n)$, and use it to make predictions in the test set $\mathcal{D}_t = (X_{n+1}, Y_{n+1}), \dots, (X_{n+m}, Y_{n+m})$. The response variables $Y_{n+1}. \dots, Y_{n+m}$ are not observed in practice, but to evaluate our method, we will use simulated data for which these outcomes are known.   

The parameter $\Theta$ is determined by the CART split criterion. We define a split in a cell $C$ is a pair $(j, s)$, where $j \in {1, \dots, p}$ denotes the splitting variable and $s$ as the potential splitting values along the \emph{j}th variable. We let $n_C$ be the number of samples falling in cell $C$. With covariates notation $X_i = (X_i^{(1)}, \dots, X_i^{(p)})$, for all possible $(j, s)$ in $C$, the CART split criterion considers maximized homogeneity in two descendant nodes \citep{scornet2015consistency}:
\begin{equation}\label{eq2}
  L_n(j, s) = - \sum_{i=1}^{n}(Y_i - \bar{Y}_{C_{l}}\mathds{1}_{X_i^{(j)} \le s} - \bar{Y}_{C_{r}}\mathds{1}_{X_i^{(j)} > s})^2\mathds{1}_{X_i \in C},
\end{equation}
where $\bar{Y}_{C_{l}} = \sum_{i \in C_l}Y_i/n_{{C_{l}}}$ and $\bar{Y}_{C_{r}} = \sum_{i \in C_r}Y_i/n_{{C_{r}}}$ are the average of $Y_i$'s in left and right descendant cells of $C$: $C_l = \{X_i \in C | X_i^{(j)} \le s\}$ and $C_r = \{X_i \in C | X_i^{(j)} > s\}$ respectively. The pair $(\hat{j}, \hat{s})$ is selected by maximizing $L_n(j, s)$ over all possible pairs $(j, s)$ in $\mathcal{D}_s$. To avoid ties in the optimization procedure, $s$ in the pair $(j, s)$ is selected in the middle of the consecutive sampling values of $X^{(j)}$. 

In practice, the CART-split criterion is implemented based on the concept of squared prediction error for regression and Gini impurity for classification \citep{hastie2009elements, therneau2023introduction}. Entropy can serve as an alternative in case of classification trees \citep{hastie2009elements}. 

It is a common assumption that the training and test data used in tree-based learning algorithms like CART are drawn independently and identically (i.i.d) from the same distribution. However, if this assumption is violated, sample selection bias can occur in real-world scenarios. In such scenarios, we refer to the training and test sets as source and target sets. First, we explain why the CART model is sensitive to sample selection bias.

\subsection{CART model under sample selection bias} 
\label{subsec2.2}
Sample selection bias is a common problem frequently ignored in machine learning algorithms. This bias occurs when the data used for analysis does not represent the entire population due to non-random sampling. Consequently, models trained on this biased data may generate inaccurate and misleading predictions when applied to new, unseen data. This happens because the training data fails to accurately capture the actual underlying patterns or distributions in the population \citep{tripepi2010selection, cortes2008sample}. 

For example, in a clinical trial to evaluate the effectiveness of a new drug for treating diabetes, patient recruitment is primarily focused on healthcare facilities and clinics. However, this recruitment strategy inherently biases the participant pool towards individuals who already have access to medical care or are receiving treatment for their condition. Consequently, the trial cohort is skewed towards individuals with mild or managed forms of diabetes while potentially excluding people with more severe or untreated manifestations.

This sample selection bias undermines the external validity of a trial by limiting its representativeness of the broader diabetic population. As a result, any conclusions drawn from the trial outcomes may not accurately generalize to the broader spectrum of diabetic patients, especially those who face difficulties in accessing healthcare or show more complicated disease presentations. Therefore, the predictive model may not be optimal, and this can compromise the reliability of subsequent clinical decisions and healthcare interventions based on the trial findings.

Let us consider a dichotomous variable $W$ that indicates the domain of the sample, where $W = 1$ signifies that the unit belongs to the target domain, and $W = 0$ indicates that the unit is in the training domain. We may address sample selection bias in classical regression models by including variables correlated with $W$ in the prediction model. When these variables are included, the outcome $Y$ becomes independent of the sample selection index $W$, ensuring that $P(Y|X, W=0) = P(Y|X, W=1) = P(Y|X)$. This method effectively mitigates sample selection bias because regression directly models the conditional probability $P(Y|X)$, remaining unaffected by the distribution of $X$. This is analogous to the ``Missing at Random" (MAR) assumption in missing data problem, where the probability of missingness depends only on the observed data \citep{little2019statistical}. 

However, sample selection bias presents a different challenge when using tree-based models. CART, as well as other tree-based models, splits the data by selecting the most informative predictor at each node, conditioning only on one variable $X^{(j)}$ during each binary split, as shown in Eq (\ref{eq2}). Since the model relies on a single variable at each split, the loss function cannot simultaneously incorporate all the variables that influence the selection mechanism. Consequently, sample selection bias persists in the split of tree-based models, meaning $L_n(j, s)$ will be different for source ($W = 0$) and target ($W = 1$) domains. The split criterion $L_n(j, s)$ is sensitive to the distribution of $X$, which influences the resulting distributions of observations in the descendant nodes. In the appendix \ref{app1}, we provide a simple example demonstrating how sample selection bias affects linear regression models and CART models differently.

In short, the distribution of split variables can affect how units are partitioned into two child nodes and the estimates in them \citep{zadrozny2004learning}. This is true regardless of whether the support of target sample is broader or more restricted.   

\section{Domain adaptive CART (DA-CART)}
\label{sec3}
\subsection{Domain adaptation with importance weighting}
Since the data used for training machine learning models may not always accurately represent the underlying distribution of the target data, developing a predictor model that can accurately predict outcomes in the target domain is useful.

When faced with an unsupervised problem, sample weighting methods can provide a practical solution. These methods involve assigning varying weights to instances in the source domain to adjust its distribution to match that of the target domain. By adjusting the influence of individual samples based on their similarity to the target domain, sample weighting methods can reduce the impact of domain shift on model performance.

Suppose the source set $\mathcal{D}_s$ consists of $n$ samples and follows a joint distribution $p'(X, Y)$. Analogously, the target set $\mathcal{D}_t$ consists of $m$ samples and follow a different joint distribution: $\mathcal{D}_t \sim p(X, Y)$. To identify importance weights, we assume $p'(X, Y)$ and $p(X, Y)$ are given by:
\begin{align}
    p'(X, Y) &= f(Y | X, W = 0)p(X | W = 0)\\
    p(X, Y) &= f(Y | X, W = 1)p(X | W = 1).
\end{align}
When only considering covariate shift, we assume that the conditional distribution remains the same in both the source and target domains: $f(Y | X, W = 0) = f(Y | X, W = 1)$. Therefore, the domain shift can be captured by the distribution discrepancy of predictors: 
\begin{equation} 
    \frac{p(X, Y)}{p'(X, Y)} =  \frac{p(X| W = 1)}{p(X| W = 0)}
\end{equation}

In general, a machine learning method minimizes the expected loss 
\begin{equation}
   \mathbb{E}_{(X,Y)\sim p'}[\ell(X, Y, \theta)] 
\end{equation}
in terms of parameters $\theta$ on the training set. The subscript $(X,Y)\sim p'$ implies the sample used for expected loss calculation follows the joint distribution $p'$. However, if the goal is to train a model on the source domain but minimize the loss function in the target domain, as we do not have $Y$ for the target domain, we have:
\begin{equation}
\begin{aligned}
 \mathbb{E}_{(X,Y)\sim p}[\ell(X, Y, \theta)] 
 = &\mathbb{E}_{(X,Y)\sim p'}[\frac{p(X, Y)}{p'(X, Y)}\ell(X, Y, \theta)] \\
 = &\mathbb{E}_{(X,Y)\sim p'}[\frac{p(X| W = 1)}{p(X| W = 0)}\ell(X, Y, \theta)],
    \end{aligned}
\end{equation}
which suggest the following estimate based on samples from $\mathcal{D}_s \sim p'(X, Y)$:
\begin{equation}\label{eq3}
    \hat{\mathbb{E}}_{(X,Y)\sim p}[\ell(X, Y, \theta)]
    = \frac{1}{n}\sum_{i=1}^{n}\frac{\hat{p}(X| W = 1)}{\hat{p}(X| W = 0)}\ell_{p'}(X_i, Y_i, \theta),
\end{equation}
where $\ell_{p'}(X_i, Y_i, \theta)$ implies that the sample used for loss function calculation follows the joint distribution $p'$.

Huang et al.~(\citeyear{huang2006correcting}) proposed kernel mean matching for importance weighting estimation. Similarly, Sugiyama et al.~(\citeyear{sugiyama2007direct}) proposed the Kullback-Leibler Importance Estimation Procedure for importance weighting. These methods are based on kernel methods, which relieve the distribution assumption of covariates. However, when dealing with high-dimensional datasets containing mixed types of variables, kernel estimates encounter several challenges. The curse of dimensionality increases computational complexity, making kernel methods impractical due to longer computation time and greater memory requirements. Additionally, mixed-type data complicates the definition of suitable kernel functions, since standard kernels assume continuous feature spaces. Handling categorical or ordinal variables within kernel methods becomes problematic without additional assumptions or discretization steps. Selecting an appropriate kernel function can be non-trivial, with numerous variables and mixed types, potentially leading to suboptimal distance measures.

Propensity score weighting, the statistical technique we apply, is a valuable tool for adjusting confounding variables in observational studies \citep{guo2014propensity, carry2021inverse}. We can effectively adjust for sample selection bias by estimating the probability of receiving a particular treatment based on observed covariates and deriving weights (propensity scores) from these estimates. These weights are then applied to each observation, upweighting or downweighting individuals based on their propensity score. In our case, we estimate the probability of being in the source and target domains. By reweighting the data in the source domain using propensity score weighting, we can assign more importance to units that are more likely to be in the target domain when determining the split. 

Propensity score weighting is often used to draw inferences about populations. The weight assigned to each observation is based on the probability of being observed, typically represented as $\boldsymbol{\omega}^{0} = 1/P (W = 0 | X)$. However, if the objective is to make targeted inferences, it follows from Eq (\ref{eq3}) and Bayes' Theorem that the importance weights need to be adjusted to $\boldsymbol{\omega}^{0} = P(W = 1 | X)/P(W = 0 | X) = c\cdot p(X | W = 1) / p(X | W = 0)$, where constant $c$ does not depend on $X$.  

The final importance weights we applied result from normalizing $\boldsymbol{\omega}$:
\begin{equation}
    \boldsymbol{\omega}_i = n\frac{\boldsymbol{\omega}_i^{0}}{\sum\boldsymbol{\omega}_i^{0}},
\end{equation}
which assumes the effective sample size remains the same $\sum_i^{n} \boldsymbol{\omega}_i = n$. The normalization implies that constant $c$ is not relevant for our procedure.

Let $\mathcal{S}$ and $\mathcal{S}'$ denote the support of $X$ in the target ($W = 1$) and source ($W = 0$) domain, respectively. We assume positivity of domain participation at the population level, that is $0 < P(W = 1 | X) < 1$. However, if a sample $X_i \in S'$ and $X_i \not\in S$, the estimated $\hat{\boldsymbol{\omega}}_i$ will be down-weighted to 0. A more delicate situation is that a simple $X_i \in S$ and $X_i \not\in S'$, which refer to extrapolation. In such a case, $\hat{\boldsymbol{\omega}}_i$ may be undefined ($\hat{\boldsymbol{\omega}}_i \rightarrow \infty$). To avoid unstable importance weights, we truncate $\hat{P}(W = 1 | X)$  to the interval $[0.05, 0.95]$ and calcuate $\hat{\boldsymbol{\omega}}_i^{0} = \hat{P}(W = 1 | X) / (1 - \hat{P}(W = 1 | X))$ before normalization. The boundary of truncate interval could be tuned. \citet{zhang2019efficient} show that importance weighting performs reasonably well when $\mathcal{S} \in \mathcal{S}'$, or when $\mathcal{S}$ shifts slightly in comparison to $\mathcal{S}'$ (minor extrapolation). We will come back to this point later in the simulation section.  

\subsection{DA-CART algorithm}
We propose the domain adaptive CART model, which comprises of three main steps:
\begin{enumerate}
    \item Select predictive variables $X_p$ in the outcome prediction model.
    \item Fit the selection model using the variables selected in step 1.
    \item Construct the DA-CART using the importance weights from step 2.
\end{enumerate}
The DA-CART is a semi-supervised CART since no outcomes are observed in the target domain. 

In step one, we analyze the global importance of variables from the source domain to determine which ones should be incorporated into the selection model, which is used to estimate importance weights. Based on \citet{brookhart2006variable}, it is advantageous to include prediction-specific variables when building the selection model. These variables are not directly linked to the selection model but are associated with the outcome prediction model. Their inclusion can improve the accuracy of estimating the probability of being in the target set without introducing bias. On the other hand, incorporating weight-specific variables, which are related to the selection model but not the outcome prediction model, can reduce the precision of the probability estimate of being in the target set without reducing bias. In smaller studies, including variables strongly associated with the selection model but weakly related to the outcome prediction model may lead to a more significant prediction error \citep{brookhart2006variable}.

To select variables for the selection model, we start by creating a CART model $\textbf{M1} = \text{CART}(\mathcal{D}_s)$ on the source domain. We define the global importance $V_i$ for $i = 1, \dots, p$, and variables with larger $V_i$ are included in the selection model. The gain method is widely used as the basis for variable selection and applied in our studies. It measures the total reduction in prediction error or impurity resulting from all splits made for a given variable \citep{sandri2008bias, huynh2010inferring}. Alternative methods include split count, increase in prediction error after permutation \citep{lundberg2018consistent}, and SHAP values \citep{lundberg2018consistent, lundberg2020local}. 

Determining which variables to include in the selection model can be challenging. Even if we have information about the importance of variables for the outcome prediction model, we still face the dilemma of whether moderately predictive variables should be included. We recommend a heuristic strategy for variable selection in which variables are chosen based on their global importance. We start with the most significant variable and continue selecting additional variables until their cumulative importance exceeds $ q\% $ of the total. In our studies, we set $q = 85\%$. However, it is crucial to consider different strategies for selecting variables based on their global importance. Additionally, background information can be considered when determining the variables in the importance weighting model.

Suppose a set of predictive variables for the outcome model $Z$ is selected for the importance weighting model $\textbf{M2}$. To improve the estimation of the importance weights, we construct an ensemble trees model (ETM): $\textbf{M2} = \text{ETM}(Z_s, Z_t, W)$, where $Z_s$ represents the selected variables from the source domain and $Z_t$ represents the same variables in the target domain. The variable $W$ indicates whether a sample is from the source or target domain and is defined in section \ref{subsec2.2}. Once $\textbf{M2}$ is constructed, we estimate the probability of being in the target domain as $\hat{P}(W=1|Z_s)$ and then calculate the normalized importance weights as $\hat{\boldsymbol{\omega}} \propto \hat{P}(W=1 | Z_s) / (1 - \hat{P}(W=1 | Z_s))$. 

Once the importance weights are estimated, a weighted CART is constructed using the predictive variables in the source domain. The model is denoted as $\textbf{M3} = \text{CART}(Z_s, \mathcal{D}_s(y), \hat{\boldsymbol{\omega}})$, with the weighted CART split criterion:
\begin{equation}
  L_n(j, s) = - \sum_{i=1}^{n}\hat{\boldsymbol{\omega}}_i(y_i - \bar{y}_{C_{l}}\mathds{1}_{X_i^{(j)} \le s} - \bar{y}_{C_{r}}\mathds{1}_{X_i^{(j)} > s})^2\mathds{1}_{X_i \in C},
\end{equation}
Overfitting is presented by pruning. Algorithm 1 depicts the pseudocode of DA-CART.\\

\begin{algorithm}[H]
\caption{DA-CART Algorithm}
\SetKwInput{KwInput}{Input}                
\SetKwInput{KwOutput}{Output}              
\DontPrintSemicolon
  
  \KwInput{$\mathcal{D}_s = \{(X_1, Y_1), \dots, (X_n, Y_n)\}$, $\{X_{n+1}, \dots, X_{n+m}\}$ from $\mathcal{D}_t$, and  $W = \{\underbrace{0, \dots, 0}_{n},\underbrace{1, \dots, 1}_{m}   
 \}$}
  \KwOutput{$\{\hat{\textbf{Y}} = \hat{Y}_{n+1}, \dots, \hat{Y}_{n+m}\}$}

  \SetKwFunction{FMain}{Prediction}
  \SetKwFunction{FSum}{Variable selection}
  \SetKwFunction{FSub}{Importance weights calculation}
 
  \SetKwProg{Fn}{Step}{:}{}
  \Fn{\FSum{$\mathcal{D}_s$}}{
        build a CART model: 
         \textbf{M1} = CART($(X_1, \dots, X_n), (Y_1, \dots, Y_n)$)\;
        calculate the variable importance of \textbf{M1}\;
        select predictive variables $Z_s$ based on the variable importance\; 
        \KwRet $Z_s$\;
  }
  \;

  \SetKwProg{Fn}{Step}{:}{}
  \Fn{\FSub{$W$, $Z_s$, $Z_t$}}{
       build an ensemble tree model (ETM) with covariates ${Z_s, Z_t}$:
       \textbf{M2} = ETM($\{Z_s, Z_t\}$, $W$)\;
    $\hat{P}(W=1 | Z_s)$ = predict(\textbf{M2}, $Z_s$)\;
    $\hat{\boldsymbol{\omega}} \propto \hat{P}(W=1 | Z_s) / (1 - \hat{P}(W=1 | Z_s))$\;
    \KwRet $\hat{\boldsymbol{\omega}}$\;
  }
  \;

  \SetKwProg{Fn}{Step}{:}{}
  \Fn{\FMain{$(Y_1, \dots, Y_n)$, $Z_s$, $Z_t$ $\hat{\boldsymbol{\omega}}$}}{
       build a weighted CART model: \textbf{M3} = CART($Z_s, (Y_1, \dots, Y_n), \hat{\boldsymbol{\omega}}$)\;
    $\hat{\textbf{Y}}$ = predict(\textbf{M3}, $Z_t$)\;
    \KwRet $\hat{\textbf{Y}}$\;
  }
\end{algorithm}

\subsection{Consistency of the estimates on the target domain}
The goal of CART is to predict a random response variable $Y$ by estimating the mean function $m_{(X,Y)\sim p'}(X) = E[Y | X = x]$. In the context of domain adaptation, the objective is to use the data $D_s$ from the source domain to create an estimate $m_n(X, \Theta') : [0, 1]^p \rightarrow R$ for the function $m_{(X, Y) \sim p}(X) = E[Y | X]$ in the target domain. Later, we will only use the subscripts $X \sim D$ and $X \sim D'$ to denote the predictors from the target or source domains, emphasizing the difference in covariate distribution. This is based on the assumption that the conditional distribution of the outcome $Y$ remains the same. The DA-CART estimate in the source domain $m_n$ is $L^2$ consistent toward the target domain if $E[m_n(X, \Theta') - m_{X\sim D}(X)]^2 \rightarrow 0$ as $n \rightarrow \infty$. 

\textbf{Theorem 1}: Suppose $n \rightarrow \infty$, $t_n \rightarrow \infty$ and $t_n(\log n)^9/n \rightarrow 0$, DA-CART $m_n(X, \Theta')$ is consistent towards the target domain, that is,
\begin{equation}
    lim_{n \rightarrow \infty}E[m_n(X, \Theta') - m_{X\sim D}(X)]^2 = 0. 
\end{equation}
We will show the proof in the appendix \ref{app2}, mainly based on the idea developed by \citep{nobel1996histogram, gyorfi2002distribution, scornet2015consistency}. In general, we connect the estimates in two domains by importance weights. The assumption in Theorem 1 implies that the depth of CART $t_n$ increases to infinity more slowly than the sample size $n$. The bound of the prediction error of CART is sufficiently controlled by $t_n$.

\subsection{Extensions of DA-CART}
\label{subsec3.4}
We extend our method by enhancing its extrapolation capabilities by incorporating the Generalized Linear Model Tree (DA-GLM-tree) \citep{rusch2013gaining, zeileis2008model}. The GLM tree approach combines decision tree partitioning with Generalized Linear Models (GLMs) at the terminal nodes. By modelling relationships beyond simple mean distributions, we expect the DA-GLM-tree to perform better than DA-CART for extrapolation.

We also extend DA-CART to bagged tree models, as ensembles of trees are known to provide more stable and reliable predictions by reducing the variability inherent in single trees \citep{breiman1996bagging, breiman2001random}. We study the performance of domain adaptation bagged trees (DA-BT), an extension of DA-CART, which incorporates importance weighting into the framework of bagged trees. We also include weighted bagged trees incorporating importance weights into the bootstrap sampling process (BT-IWB). 

\subsection{Variations of important weights estimation}
\label{subsec3.5}
We investigate two variations of our method that omit the first step of DA-CART. The first variation, DA-CART-naive, uses all covariates $X_1, \ldots, X_p$ in the selection model, regardless of their relevance for the outcome prediction model. DA-CART-naive may lead to a worse performance when the actual selection model includes variables that do not contribute to the outcome prediction model. This is because the importance weights should not rely on variables not predictive of the outcome prediction model \citep{brookhart2006variable}. The second variation, DA-CART-kernel, estimates kernel-based importance weights using Kullback-Leibler distance \citep{sugiyama2007direct}.

We also extend the DA-GLM-tree to two variations, DA-GLM-tree-naive and DA-GLM-tree-kernel, following a similar approach to extend DA-CART to DA-CART-naive and DA-CART-kernel. All extensions in subsection \ref{subsec3.4} and variations in subsection \ref{subsec3.5} are evaluated through simulation studies.

\section{Simulation study}
\label{sec4}
\subsection{Data generating model}\label{subsec4.1}
In our study, we conducted a series of simulations to evaluate the performance of DA-CART. We generated data from the following model \citep{friedman1991multivariate}:
\begin{equation}
y = 5\text{sin}(X_{1}X_{2}) + X_{1} + X_{1}^2 + X_{2} + X_{2}^2 + \epsilon,    
\end{equation}
where $\epsilon$ follows a standard normal distribution. We generated data for the entire population using five independent predictors: $X_1 \sim \text{N}(0, 3)$, $X_2 \sim \text{N}(0, 1)$, $X_3 \sim \text{Uni}(0, 1)$, $X_4 \sim \text{N}(0, 1)$, and $X_5 \sim \text{Gamma}(2, 1)$, where$X_3, X_4$ and $X_5$ are noise variables. 

The sub-populations for the source and target domains are derived from an overall population using a specified selection model. We first defined a score variable within this selection model. The score variable is a linear combination of covariates, similar to the method used for simulating missing values \citep{schouten2018generating}. We developed two linear combinations for the score variable: \(\texttt{score} = X_1\) and \(\texttt{score} = X_1 + 2X_4\).

Additionally, we defined two patterns for the selection model: restricted range and shifted range. For the restricted range pattern, the selection model is \(P(W = 1) = \text{logit}(-|\texttt{score} - \overline{\texttt{score}}| + 2)\). A unit is assigned to the target domain if \(W=1\); otherwise, it is assigned to the source domain. The selection model for the restricted range indicates that units with scores closer to the centre are more likely to be sampled in the target domain.

In the shifted range scenario, the selection model is defined as \(P(W = 1) = \text{logit}(\texttt{score} - \overline{\texttt{score}})\). This means that units with higher score values are more likely to be sampled in the target domain. The restricted range scenario is related to an interpolation problem, whereas the shifted range scenario corresponds to an extrapolation problem, which is more challenging for domain adaptation.

We used a CART model and the gain method to select predictive variables from the outcome model in step 1 of DA-CART. Variables are ranked in descending order of information gain. Variables are ranked in descending order based on information gain.  The selection begins with the most significant variable, and additional variables are included sequentially until their cumulative importance exceeds 85\% of the total. 
The XGBoost model was applied to estimate the selection model, which was then used to calculate the importance weights. The sample size of the test set is fixed to 10000 in the target domain. The training set could be from the source or target domain.  

In short, we varied the following factors systematically:
\begin{enumerate}
    \item patterns of covariate shift: restricted range or shifted range.
    \item variables in the selection model: $X_1$ or $X_1$ and $X_4$.
    \item sample size of the training set: $n = \{500, 1000, 5000, 10000\}$.
\end{enumerate}
We denote the variables in the selection model as sel, while the variables in the outcome prediction model are referred to as pred. Four distinct scenarios, resulting from variations in the first two factors listed above, are used as a faceting variable in the resulting visualization:
\begin{enumerate}
    \item Restricted range with sel $\subseteq$ pred: restricted range with $\texttt{score} = X_1$, where \textbf{all} variables in the selection model are included in the outcome prediction model.
    \item Restricted range with sel $\subsetneq$ pred: restricted range with $\texttt{score} = X_1 + 2X_4$, where \textbf{not all} variables in the selection model are included in the outcome prediction model.
    \item Shifted range with sel $\subseteq$ pred: shifted range with $\texttt{score} = X_1$, where \textbf{all} variables in the selection model are included in the outcome prediction model.
    \item Shifted range with sel $\subsetneq$ pred: shifted range with $\texttt{score} = X_1$, where \textbf{not all} variables in the selection model are included in the outcome prediction model.
\end{enumerate}
The distribution discrepancies of $X_1$ and $X_4$ under different score variables and selection models are visualized in the appendix \ref{app3}.

\subsection{Compared methods}
We compared DA-CART with two baseline CART models: a naive CART model trained on the source domain (Naive CART) and a target CART model trained on the target domain (Target CART). The Naive CART model represents the standard approach typically used in practice. In contrast, the Target CART model serves as a benchmark for performance evaluation, although it is usually unavailable in real-world scenarios.

We also evaluated a weighted CART model using oracle weights (weighted CART with OW), which are the weights from the simulation. Additionally, we considered two domain-adaptive variations of DA-CART: one is a naive weighting approach (DA-CART-naive), while the other employs a kernel-based weighting strategy (DA-CART-kernel).

To expand our comparison, we included the DA-GLM-tree and its variations: DA-GLM-tree-naive and DA-GLM-tree-kernel. Furthermore, we assessed three other GLM tree models: a Naive GLM tree, a Target GLM tree, and a weighted GLM tree that uses oracle weights (weighted GLM tree with OW). The Naive GLM tree and Target GLM tree serve as generalizations of their corresponding CART counterparts within the GLM tree framework.

\subsection{Results}
\subsubsection{Performance under covariate shift}
\begin{figure}[H]
    \centering
    \includegraphics[width=\textwidth, height=0.5\textheight]{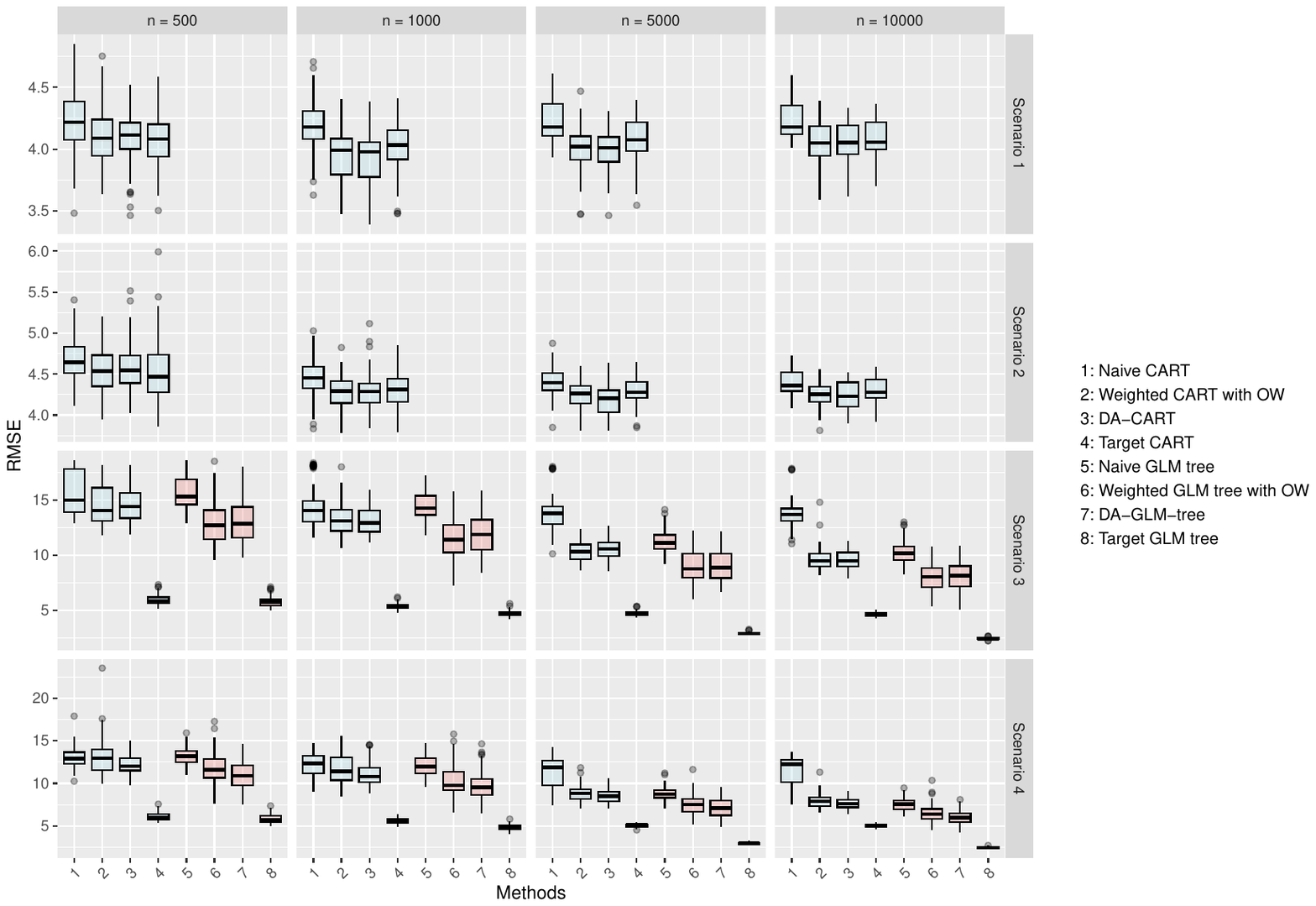}
    \caption{This figure presents boxplots of Root Mean Square Error (RMSE) across various tree models based on 100 repeated simulations. It consists of a 4 × 4 panel grid organized into columns representing different sample sizes (500, 1,000, 5,000, and 10,000) and rows for various scenarios. The scenarios are as follows: 1. Restricted range with sel $\subseteq$ pred 2. Restricted range with sel $\subsetneq$ pred 3. Shifted range with sel $\subseteq$ pred 4. Shifted range with sel $\subsetneq$ pred. All methods being compared are labelled on the x-axis, and the corresponding model names are listed in the legend.}
    \label{fig1}
\end{figure}

We evaluated the performance of DA-CART under both restricted range and shifted range conditions. The prediction errors are illustrated in Fig. \ref{fig1}. The top two rows display the results under the restricted range condition. Regardless of whether all variables from the selection model are included in the outcome prediction model, DA-CART consistently outperforms the Naive CART model and achieves a root mean square error close to that of the Target CART. When sel $\subsetneq$ pred, DA-CART performs better than the weighted CART model, particularly as the sample size increases. This emphasizes the necessity of step 1 in the DA-CART method and supports the validity of Theorem 1.

We also assessed DA-CART's performance under the shifted range condition, which challenges tree-based models and importance weights estimation, as noted by \citet{zhang2019efficient}. Our findings indicate that DA-CART is less effective in this context than target CART, representing optimal performance but only achievable in practice when target outcomes are available. However, as the sample size increases, DA-CART's performance improves due to the increased likelihood of having representative target domain units in the source domain.

As the GLM tree improves extrapolation capabilities, the DA-GLM-tree outperforms the DA-CART in shifted range situations. The proposed domain adaptation strategy significantly improves the predictive accuracy of the Naive GLM tree, demonstrating the broad applicability of our approach.

The performance of the variations of DA-CART and DA-GLM-trees (DA-CART-naive, DA-CART-kernel, DA-GLM-tree-naive, and DA-GLM-tree-kernel) is presented in the appendix \ref{app4}. Overall, DA-CART and DA-GLM-trees outperform their extensions.

\subsubsection{Performance in non-discrepancy scenarios}
We also explore the performance of DA-CART when the source and target sets are drawn from the same data generating model, as introduced in section \ref{subsec4.1}, with $P(W = 1) = 0.5$. In this scenario, estimating the importance weights does not worsen prediction performance. The resulting plot is in the appendix \ref{app5}, as shown in Fig. \ref{fig11}.

\subsubsection{Importance weighting in bagged tree (BT) models}
In this manuscript, we primarily focus on DA-CART. We also explore how domain adaptation techniques can be used with tree ensembles, specifically bagged trees. We study the performance of DA-BT, which is an extension of DA-CART. It applies importance weights during tree splitting to ensure that the splits accurately reflect the characteristics of the target domain. Our analysis focuses on sampling uncertainty, while we do not consider variable selection uncertainty across multiple trees. It allows us to examine domain adaptation's impact on model performance within tree ensembles.

\begin{figure}[H]
    \centering
    \includegraphics[width=\textwidth]{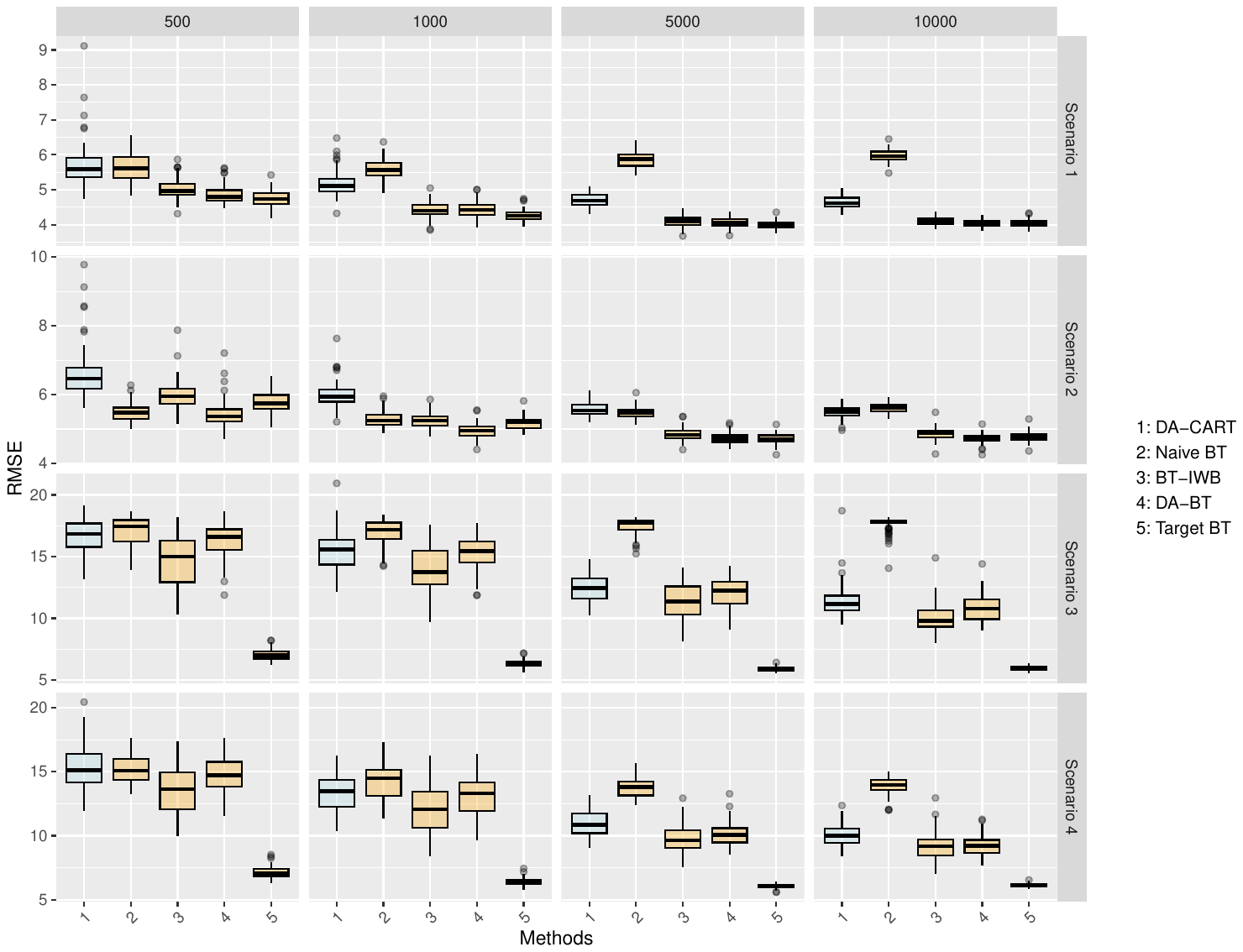}
    \caption{This figure presents boxplots of Root Mean Square Error (RMSE) across various tree models based on 100 repeated simulations. It consists of a 4 × 4 panel grid organized into columns representing different sample sizes (500, 1,000, 5,000, and 10,000) and rows for various scenarios. The scenarios are as follows: 1. Restricted range with sel $\subseteq$ pred 2. Restricted range with sel $\subsetneq$ pred 3. Shifted range with sel $\subseteq$ pred 4. Shifted range with sel $\subsetneq$ pred. All methods being compared are labelled on the x-axis, and the corresponding model names are listed in the legend.}
    \label{fig2}
\end{figure}

The details of the simulation settings, such as the data-generating model, selection methods, and sample sizes, can be found in Section \ref{subsec4.1}. The result is shown in Fig. \ref{fig2}. In these simulations, we compare DA-BT with various models, including DA-CART. The baseline model, Naive bagged trees (BT), uses standard bootstrapping without considering domain shifts. The target BT model, trained only on target domain data, represents the ideal scenario where outcomes in the target domain are observed. The comparison also includes BT-IWB, which uses oracle weights as importance weights..

Regarding performance, it has been observed that naive bagged trees consistently produce inferior results, especially as the sample sizes increase. This emphasizes the advantage of domain-adaptive models such as DA-CART when significant variations exist between the distributions of the source and target domains. In scenarios with restricted ranges, both importance weighting methods (BT-IWB and DA-BT) perform similarly to the target BT as the sample sizes increase, showcasing their ability to adapt effectively to the target domain. However, this convergence is less common in scenarios with shifted range, where the differences between the domains are more pronounced.

In comparing the BT-IWB and DA-BT, it has been observed that BT-IWB generally outperforms DA-BT, except in cases with restricted ranges and when the selection model is $\texttt{score} = X_1 + 2 * X_4$. However, since the uncertainty of variable selection across multiple trees is not considered, which domain-adaptive strategy is definitively superior still needs to be determined.

\section{Application}
\label{sec5}
We apply our method to the prognosis of Diffuse large B-cell lymphoma (DLBCL) patients to show the performance of DA-CART for classification. DLBCL is a common non-Hodgkin lymphoma for which accurate prognosis is challenging because of the patient's clinical and biological heterogeneity. The distributions of predictors vary in various medical institutions. Furthermore, since the days of overall survival (OS) and progression-free survival (PFS) often take several years to track, it is time-consuming for a medical institution to collect enough samples to build a prediction model. A reasonable strategy is to create the model on the source with sufficient follow-up, considering the importance weighting to adapt to the target. 

One widely recognised prognostic clinical covariate is the international prognostic index (IPI), which scores patients based on their age, the stage of the tumour, lactose dehydrogenase levels, a mobility measure, and the number of extranodal sites. We did not include copy number variations and mutations as variables in the model, as according to \citep{goedhart2023co}, these two groups of variables do not contribute to the classification. Therefore, we only include clinical and translocation variables in the model. Additionally, five variables that measure different aspects of IPI were also used to allow more flexibility. 

\begin{figure}
 \centering
     \begin{subfigure}[b]{0.4\textwidth}
         \centering
         \includegraphics[width=\textwidth]{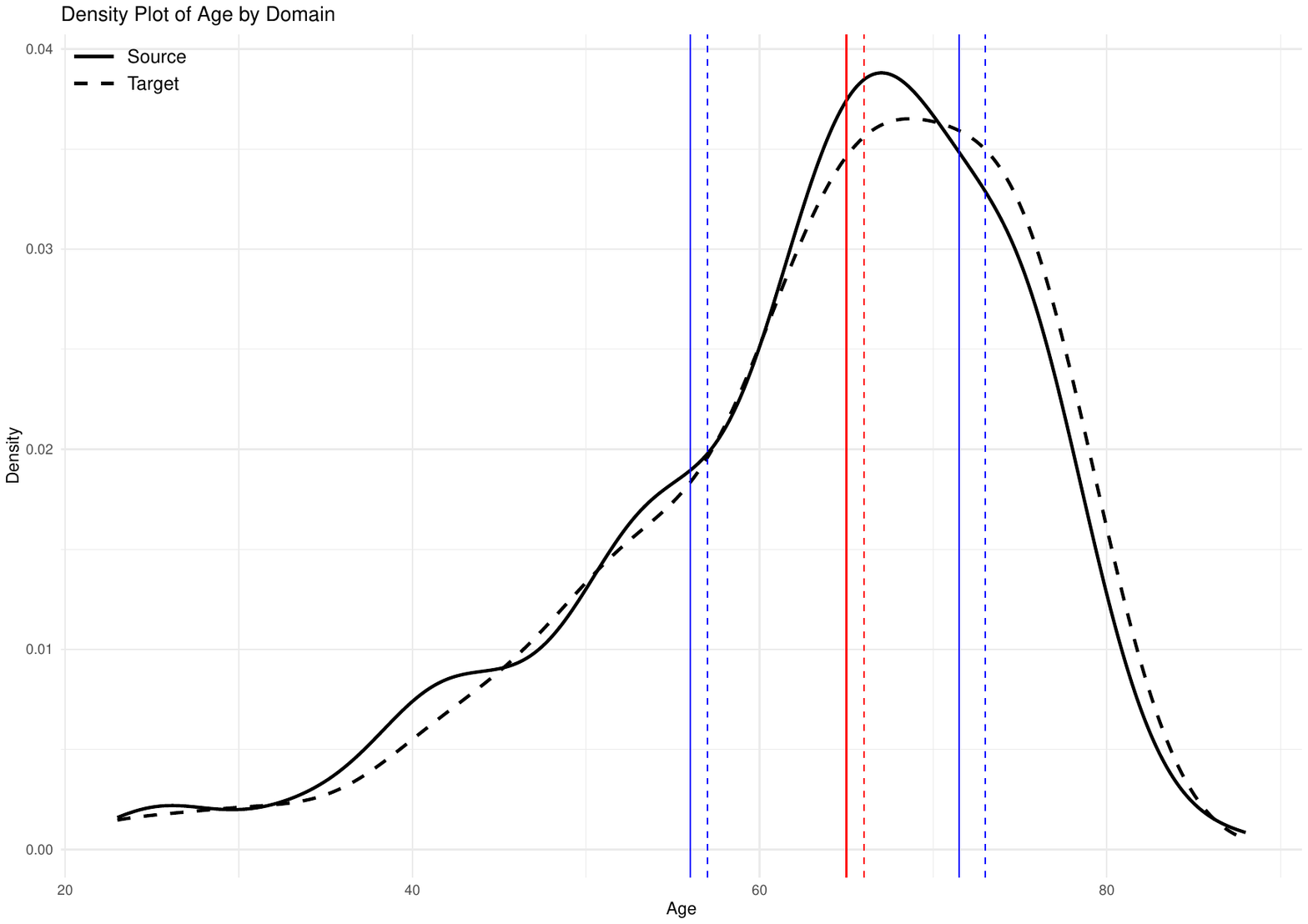}
         \caption{Age}
         \label{fig3a}
     \end{subfigure}
     \begin{subfigure}[b]{0.4\textwidth}
         \centering
         \includegraphics[width=\textwidth]{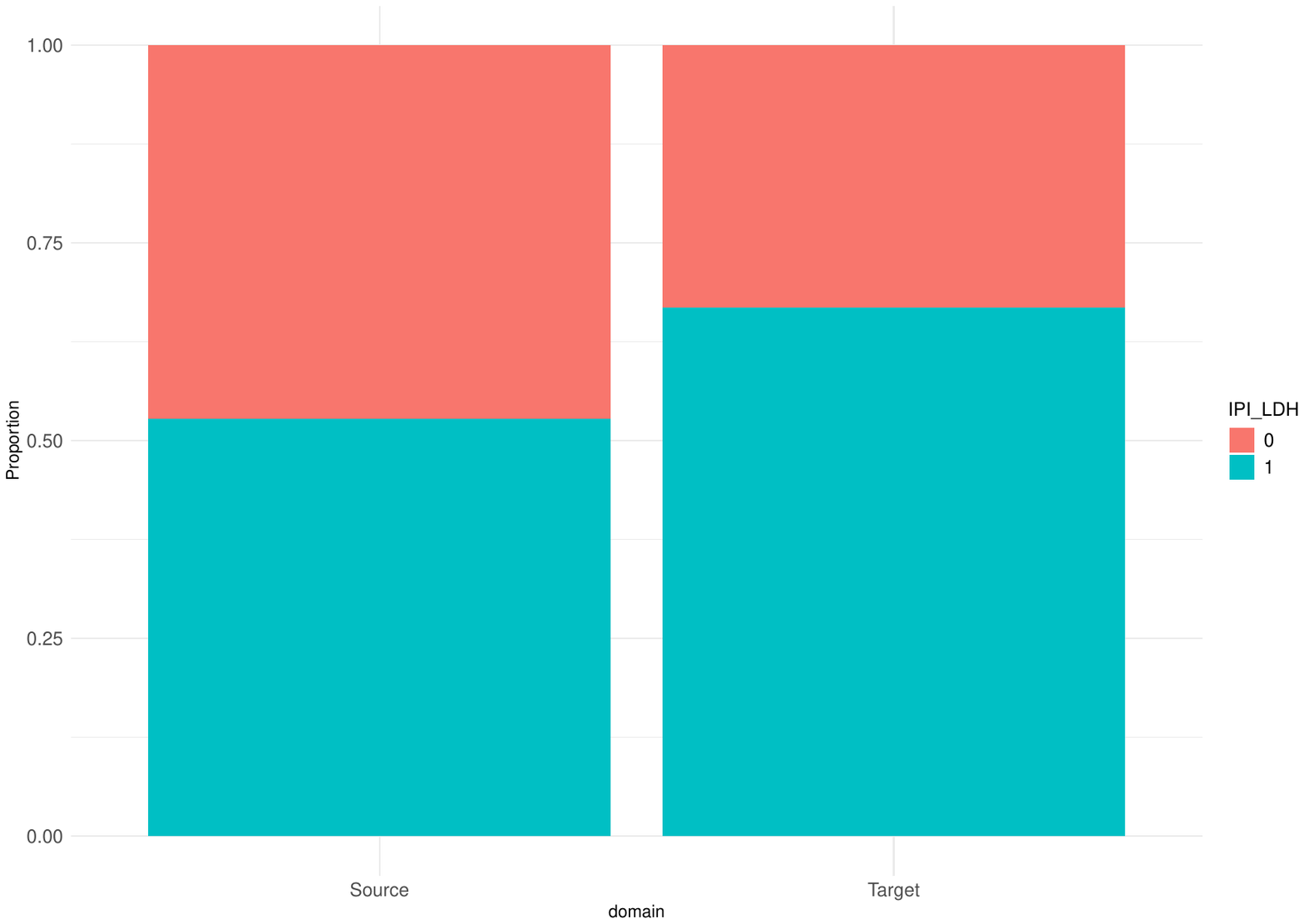}
         \caption{IPI\textunderscore LDH}
         \label{fig3b}
     \end{subfigure}
     \caption{Distribution discrepancy. For density plot of age, vertical lines indicate the 25th percentile, median, and 75th percentile for each domain.}
        \label{fig3}
\end{figure}

We fit a DA-CART model based on cohorts PETAL, Chapuy and HO130 with 415 patients to predict two-year progression-free survival in cohort HO84 with 190 patients. The data is partitioned to address the apparent distribution discrepancy between the source and target domains, see Fig. \ref{fig3}. We treat the outcome as binary because two years is a clinically well-accepted cut-off, and censoring was absent within this period.

The selection model includes the variables "Age", IPI\textunderscore LDH", and "transMYC", which together account for 86.5\% of the total variable importance. We conducted a comparison between DA-CART and naive CART. The AUC value for DA-CART is 0.655, whereas for naive CART, it is 0.612. The difference between the AUC values is nearly significant, as checked using a one-sided DeLong's pair test (p = 0.059) \citep{delong1988comparing}. Although the covariate shift is moderate, as shown in Fig. \ref{fig3}, the improvement in AUC is noticeable.

We also compared DA-CART and target CART, which trains and predicts with 5-fold cross-validation. The AUC value for the target CART is 0.66, showing no significant improvement compared to DA-CART (0.660 vs 0.655), even though the former uses the actual target labels.

 \begin{figure}
 \centering
     \begin{subfigure}[b]{0.4\textwidth}
         \centering
         \includegraphics[width=\textwidth]{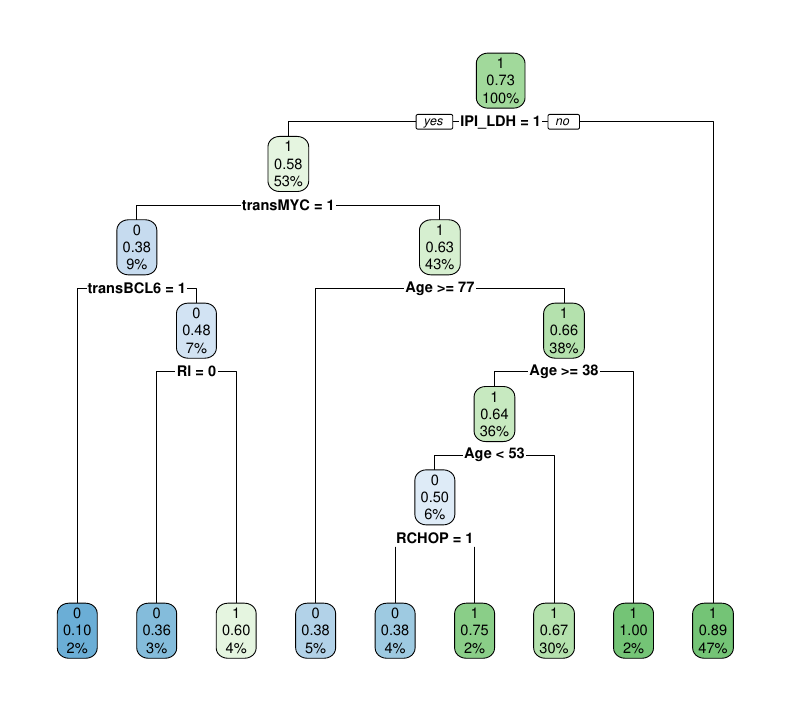}
         \caption{Naive CART}
         \label{fig4a}
     \end{subfigure}
     \begin{subfigure}[b]{0.4\textwidth}
         \centering
         \includegraphics[width=\textwidth]{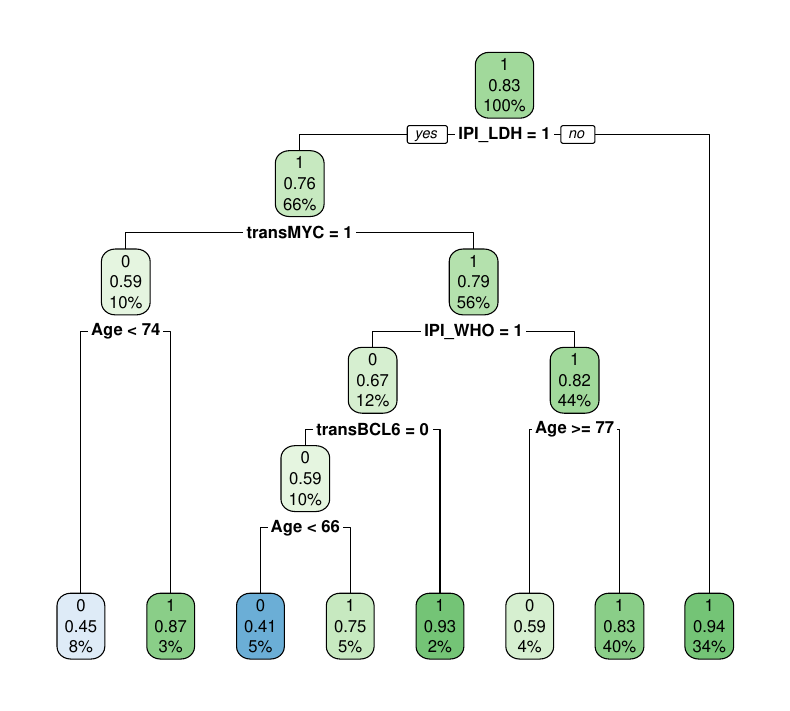}
         \caption{DA-CART}
         \label{fig4b}
     \end{subfigure}
     \caption{Visualization of naive CART and DA-CART.}
        \label{fig4}
\end{figure}

\begin{figure}
 \centering
     \begin{subfigure}[b]{0.4\textwidth}
         \centering
         \includegraphics[width=\textwidth]{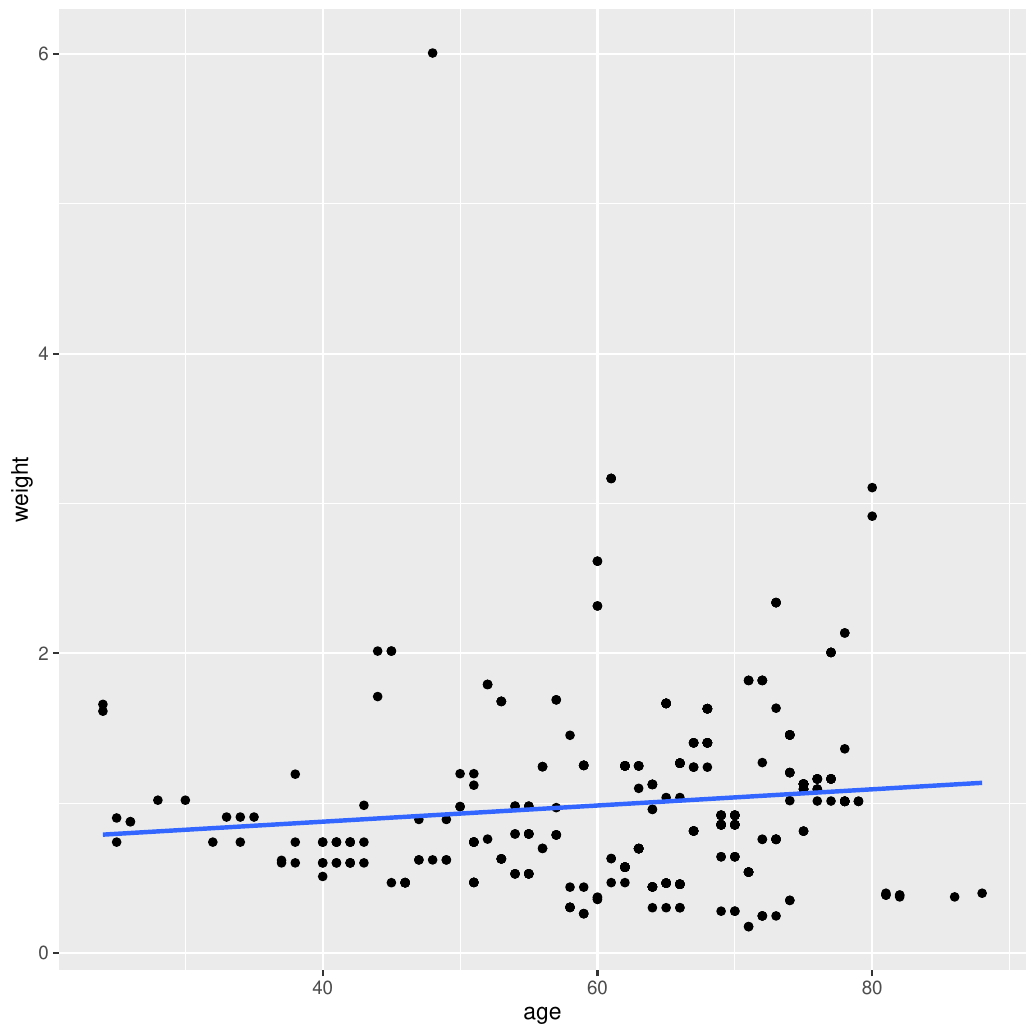}
         \caption{age vs. weight}
         \label{fig5a}
     \end{subfigure}
     \begin{subfigure}[b]{0.4\textwidth}
         \centering
         \includegraphics[width=\textwidth]{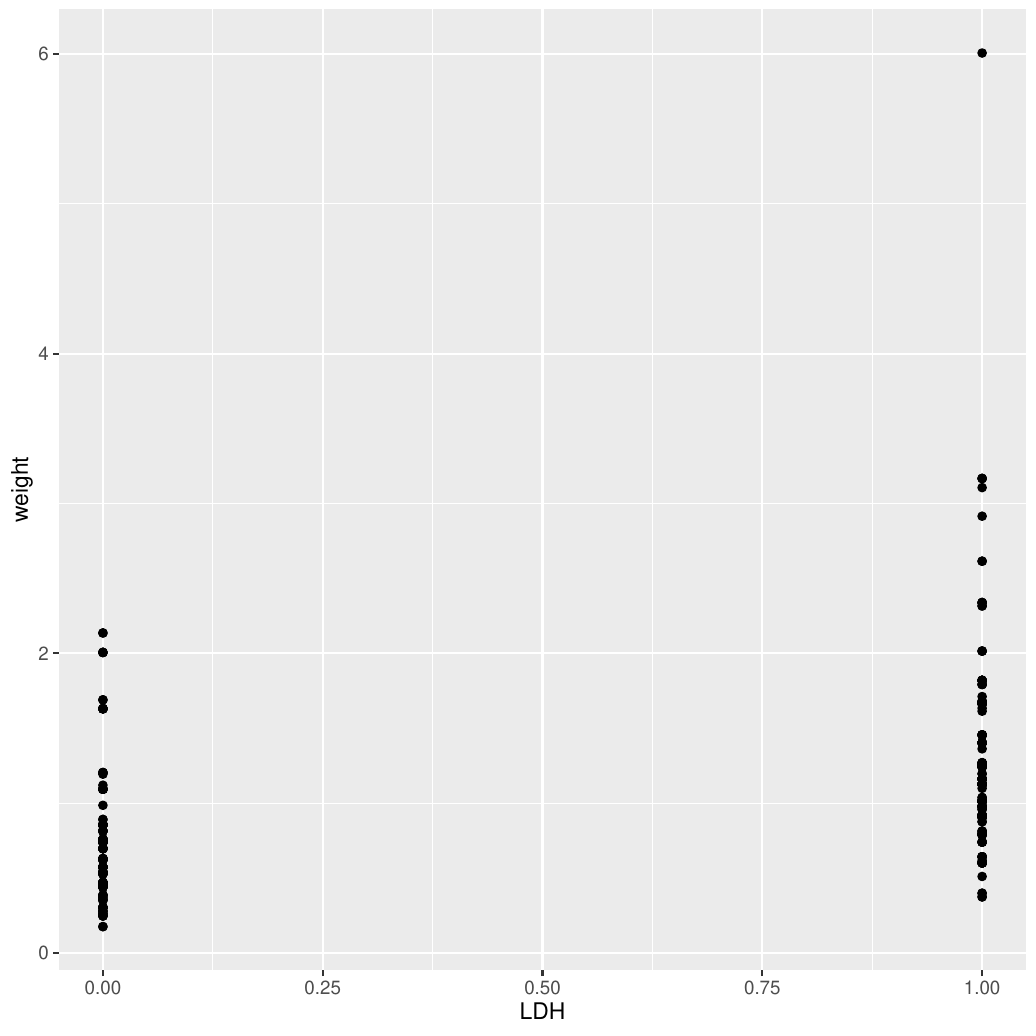}
         \caption{IPI\textunderscore LDH vs. weight}
         \label{fig:5b}
     \end{subfigure}
     \caption{How age and IPI\textunderscore LDH vs. weight affect importance weights.}
        \label{fig5}
\end{figure}

Fig. \ref{fig4} illustrates how the importance weighting affects the proportion estimates and splits. The split value remains constant for binary variables at 0.5, while the proportion changes in child nodes. The split $\text{Age}<53$ in naive CART and  $\text{Age}<66$ in DA-CART reflects a shift in the mean towards the right (the mean of Age in the target domain is higher than in the source domain). Fig. \ref{fig5} depicts how the importance weighting reflects the distribution discrepancy. Since the variable Age shifts towards the right, the importance weights are larger for older patients. Similarly, as the proportion $P_{\text{target}}(\text{IPI\textunderscore LDH} = 1) > P_{\text{source}}(\text{IPI\textunderscore LDH} = 1)$, the importance weights are larger for source samples where $\text{IPI\textunderscore LDH} = 1$.

\section{Discussion}
\label{sec6}
We developed a new method called DA-CART, which improves the accuracy of predictions when the strong predictive variables have different distributions in the source and target domains. This is achieved by incorporating importance weighting into the CART algorithm. We proved the consistency of DA-CART, which was also demonstrated in simulations. The method is particularly effective when the target dataset has a restricted range or a moderately shifted range relative to the source dataset. When there is a substantial shifted range between the distributions, the predictions are inherently degraded, although we have shown that the extension to GLM-tree improves extrapolation. In practice, it is generally reasonable to assume that the source and target domain share similar support. Moreover, as the source sample size grows, the probability of samples from the target appearing in the source set increases despite potential differences in their covariate distributions.

This manuscript focuses on CART, a single-tree model essential to random forests. DA-CART can be extended to tree ensembles by incorporating importance weighting with bagging techniques, involving specifying sampling weights with importance weighting for bootstrap or incorporating importance weights into the splitting criterion. Simulations on bagged trees explored the incorporation of importance weights to both the splitting criterion and the bootstrap sample weights. The results indicated that using weights for bootstrap samples slightly improves prediction performance in our simulation settings. However, depending on the methods and research questions, integrating weights into tree models may be done in various ways. There are indeed some examples that incorporate weights into the splitting criterion, such as gradient descent optimization algorithms \citep{ruder2016overview} and the Generalized Random Forests \citep{athey2019generalized}.

For future research, investigating semi-supervised learning scenarios, where only a small subset of samples is labeled while the majority remains unlabeled, could prove valuable. A critical challenge is enhancing a model developed in the source domain to effectively utilize limited labeled data and abundant unlabeled data in the target domain. There are opportunities to explore domain adaptation in non-adaptive tree models, such as Bayesian Additive Regression Trees (BART). Unlike adaptive models that utilize optimization criteria for splits, it is challenging for BART to transfer knowledge across domains with varying data distributions. Integrating importance weighting into the Bayesian framework for tree-based models, particularly by moderating priors instead of the likelihood to adapt from a source to a target, presents an exciting avenue for exploration. There is still an open question of whether one should tune hyperparameters of tree models (e.g., using cross-validation) when using importance weighting. This problem arises because the C.V. sets from the source may not have the same covariate distribution as the target set. This manuscript used the default hyperparameters provided in R packages for all models.

\appendix
\section{Performance comparison under sample selection bias}
\label{app1}
In order to compare the performance of a linear regression model and a CART model with sample selection bias for the source and target domains, we conducted a simple example. The dataset included 2000 samples with three standard normally distributed covariates: $(X_1, X_2, X_3 \sim N(0, 1)$. The outcome variable $Y$ was created as a linear combination of the covariates $Y = 3X_1 + 2X_2 + 0.5X_3 + \epsilon$, where $\epsilon \sim N(0, 1)$ represents standard normally distributed error terms. The probability of a sample being assigned to the source domain was determined using a logistic function based on $X_2$: $(P(W = 1 | X_2) = \frac{1}{1 + e^{-2x_2}}$. Based on this probability, samples were then split into source $(W = 1)$ and target $(W = 0)$ domains. We fitted a linear regression and a CART model to the source domain and evaluated the Mean Squared Error on the target domain. The simulation was repeated 100 times, resulting in a mean MSE of 1.01 for the linear regression model and 5.55 for the CART model.

\section{Proof of Theorem 1}
\label{app2}
\textbf{Notations}: The partition derived from the data set $D_n$ is denoted by $P_n$. We define $M(P_n)$ as the number of cells among the partition $P_n$. Given a dataset $X = \{X_1, \dots, X_n\ \subset [0, 1]^d\}$, $\Gamma(X, P_n)$ denotes the number of distinct partitions of $X$. 

Let $(\beta_n)_n$ be a positive sequence, and define the truncated operator $T_{\beta_n}$ by 
\begin{align*}
    T_{\beta_n}u &= u, if |u| < \beta_n \\
    T_{\beta_n}u &= sign(u)\beta_n, if |u| \geq \beta_n.
\end{align*}
We further define $Y_L = T_LY$ and $Y_{i, L} = T_LY_i$.

To prove Theorem 1, we need to first work out the Theorem 10.2 in \citep{gyorfi2002distribution} for DA-CART.

\textbf{Theorem 2}: Assume that:
\begin{enumerate}
    \item $\text{lim}_{n \rightarrow \infty} \beta_n = \infty$
    \item $\text{lim}_{n \rightarrow \infty}E[\text{inf}_{||f||_{\infty}\leq \beta_{n}, f\in \mathcal{F}(\Theta')}\; E_{X\sim D'}[f(X)-m_{X\sim D}(X)]^2]=0$
    \item for all $L > 0$,
          $\text{lim}\;E_{X\sim D'}[\text{sup}_{||f||_{\infty}\leq \beta_{n}, f\in \mathcal{F}(\Theta')}|\frac{1}{n}\sum_{i=1}^{n} \boldsymbol{\omega}_i[f(X_i) - Y_{i, L}]^2 - E_{X\sim D}[f(X) - Y_L]^2|] = 0$
\end{enumerate}
Then 
\begin{equation}
    \text{lim}\;E[T_{\beta_n}m_{n}(X, \Theta') - m_{X\sim D}(X)]^2 = 0
\end{equation}

We will begin by implementing the proof of theorem 2 in the source domain.

\textbf{Statement 1}: let $\beta_n = ||m_{X\sim D'}||_{\infty} + \sigma\sqrt{2}(\log(n))^2$, then $\text{lim}_{n \rightarrow \infty} \beta_n = \infty$.

\textbf{Statement 2}: Let $f_{n, \Theta'} = \sum_{C \in P_n} m_n(Z_C)\mathds{1}_C$, where $z_C \in C$ is an arbitrary point of the in cell $C$. Because $||m_n||_{\infty} < \infty$, and for large enough $n$, $\beta_{n} > ||m_n||_{\infty}$, we have:
\begin{align*}
    &E\; \text{inf}_{||f||_{\infty}\leq \beta_{n}, f\in \mathcal{F}(\Theta')}\; E_{X\sim D'}[f(X) - m_{X\sim D}(X)]^2 \\ 
    =& E\; \text{inf}_{||f||_{\infty}\leq \beta_{n}, f\in \mathcal{F}(\Theta')}\; E_{X\sim D'}[f(X) - m_{X\sim D'}(X)]^2\\
    \le &E\; \text{inf}_{||f||\leq ||m_n||_{\infty}, f\in \mathcal{F}(\Theta')}\; E_{X\sim D'}[f(X) - m_{X\sim D'}(X)]^2\\
    \le &E[m_n(Z_{C_n(X, \Theta')})-m_{X\sim D'}(X)]^2\\
    \le & E[\Delta(m_n, C_n(X, \Theta'))]^2\\
    \le & \xi^2 + 4||m_n||_{\infty}^2P[\Delta(m_n, C_n(X, \Theta')) > \xi]
 \end{align*}
where $C_n(X)$ is the cell of the DA-CART built in the source domain that contains $X$, and $\Delta(m_n, C_n(X, \Theta'))$ is the variation of $m_n$ within cell $C_n(X)$:
\begin{equation*}
    \Delta(m_n, C_n(X, \Theta')) = sup|m_n(x, \Theta') - m_n(x', \Theta')|:x,x' \in C_n(X, \Theta').
\end{equation*}
Based on the proposition 2 in \citep{scornet2015consistency}, for large enough $n$,
$E\; \text{inf}_{||f||_{\infty}\leq \beta_{n}, f\in \mathcal{F}(\Theta')}\; E_{X\sim D'}[f(X) - m_{X\sim D}(X)]^2 < 2\xi^2$, which concludes statement 2.

\textbf{Statement 3}: fix $L>0$,Then for all $n$ large enough such that $L < \beta_n$,
\begin{align*}
    &P(\text{sup}_{||f||_{\infty}\leq \beta_{n}, f\in \mathcal{F}(\Theta')}|\frac{1}{n}\sum \boldsymbol{\omega}_i[f(X_i) - Y_{i, L}]^2 - E_{X\sim D}[f(X) - Y_L]^2| > \xi) \\
    =&P_{X \sim D}(\text{sup}_{||f||_{\infty}\leq \beta_{n}, f\in \mathcal{F}(\Theta)} |\frac{1}{m} \sum[(f(X_i) - Y_{i, L})^2] - E[f(X) - Y_L]^2 | >\xi). \\
\end{align*}

For statement 3, we will focus on the target domain. Based on equation Eq (\ref{eq1}), Eq (\ref{eq3}), and the formula of $\boldsymbol{\omega}_i$, we can obtain $m_n(X, \Theta') \equiv m_m(X, \Theta)$, where: 
\begin{equation}
    \begin{aligned} 
    &m_n(X, \Theta') := \underset{f\in \mathcal{F}(\Theta')}{\argmin} \frac{1}{n} \sum \boldsymbol{\omega}_i |f(X_i) - Y_i|^2,\\
   & m_m(X, \Theta) := \underset{f\in \mathcal{F}(\Theta)}{\argmin} \frac{1}{m} \sum |f(X_i) - Y_i|^2.
\end{aligned}
\end{equation}
The equivalence implies that two models are expected to produce the same predictions. Here, $m_m(X, \Theta)$ represents an estimate of $m_{X\sim D}(X)$ based on an imaginary data set $D_t$ with sample size $m$ in the target domain where we can derive an estimate $m_m$ the same as $m_n$. In reality, $m$ could be treated as the effective sample size related to $n$:
\begin{equation}
    m = \frac{Var_{X\sim D'}(\boldsymbol{\omega}_i |f(X_i) - Y_i|^2)}{Var_{X\sim D}(|f(X_i) - Y_i|^2)}n.
\end{equation}

\begin{equation*}
\begin{aligned}
    P_{X \sim D}(&\text{sup}_{||f||_{\infty}\leq \beta_{n}, f\in \mathcal{F}(\Theta)} |\frac{1}{m} \sum[(f(X_i) - Y_{i, L})^2] - E[f(X) - Y_L]^2 | >\xi)\\
    \le  8&\text{exp}(-\frac{mK}{\beta_m^4}),
\end{aligned}
\end{equation*}
where $K \rightarrow \frac{\xi^2}{2048}$ as $m \rightarrow \infty$ \citep{scornet2015consistency}.
Finally, since
\begin{equation*}
    \text{sup}_{||f||_{\infty}\leq \beta_{n}, f\in \mathcal{F}(\Theta)} |\frac{1}{m}\sum[(f_{X\sim D}(X_i) - Y_{i, L})^2] - E[f_{X\sim D}(X) - Y_L]^2 | < 2(\beta_{m} + L)^2,
\end{equation*}
for large enough $m$, 
\begin{align*}
    E_{X\sim D}[&\text{sup}_{||f||_{\infty}\leq \beta_{n}, f\in \mathcal{F}(\Theta)} |\frac{1}{m}\sum[(f(X_i) - Y_{i, L})^2] - E[f(X) - Y_L]^2 |] \\
    &\leq \xi + 16(\beta_{m} + L)^2exp(-\frac{mK}{\beta_m^4})\\
    &\leq 2\xi,
\end{align*}
which concludes statement 3. Hence
\begin{equation*}
    \text{lim}\;E[T_{\beta_n}m_{n}(X) - m_{X\sim D}(X)]^2 = 0
\end{equation*}

For untruncated estimates:
\begin{align*}
    E[m_{n}(X, \Theta') - m_{X\sim D}(X)]^2 &= E[E_{X\sim D'}[m_n(X, \Theta')] - m_{X\sim D}(X)]^2\\
    &(E_{X\sim D'}[m_n(X, \Theta')] \text{expectation over $\Theta'$})\\
    &\leq E[m_{n}(X, \Theta') - T_{\beta_{n}}m_{n}(X, \Theta')]^2\\
        &+ E[T_{\beta_{n}}m_{n}(X, \Theta') - m_{X\sim D'}(X)]^2\\ 
    & \leq E[m_{n}^2(X, \Theta')\mathds{1}_{m_{n}(X, \Theta') \geq \beta_{n}}] + \xi    
\end{align*}
Since $|m_{n}(X, \Theta') \leq ||m_n||_{\infty} + \text{max}_{1\leq i \leq n} |\xi_i|$,
\begin{align*}
   &E[m_{n}^2(X, \Theta')\mathds{1}_{m_{n}(X, \Theta') \geq \beta_{n}}]\\
   \leq &E[(2|| m_{n} ||_{\infty}^2 + 2 \text{max}_{1\leq i \leq n} \xi_i^2)\mathds{1}_{max_{1 \leq i \leq n} \xi_i \geq \sigma\sqrt{2}log(n)^2}]\\
   \leq & 2||m_n||_{\infty}^2P[\text{max}_{1 \leq i \leq n} \xi_i \geq \sigma\sqrt{2}log(n)^2]\\
   &+2(E[\text{max}_{1 \leq i \leq n}\xi_i^4]P[\text{max}_{1 \leq i \leq n} \xi_i \geq \sigma\sqrt{2}log(n)^2])^{1/2}.
\end{align*}
Because
\begin{align*}
    P[\text{max}_{1 \leq i \leq n} \xi_i \geq \sigma\sqrt{2}log(n)^2] \leq \frac{n^{1-\log(n)}}{2\sqrt{\pi}(\log(n))^2},
\end{align*}
and $\xi_i$'s are normal distributed with mean 0. For $n \rightarrow \infty$,
\begin{align*}
    E[m_{n}(X, \Theta') - &m_{X\sim D'}(X)]^2 \\
    &\leq 2|| m_{n} ||_{\infty}^2\frac{n^{1-\log(n)}}{2\sqrt{\pi}(\log(n))^2} + \xi + 2(3n\sigma^4\frac{n^{1-\log(n)}}{2\sqrt{\pi}(\log(n))^2})^{1/2}\\
    & \leq 3\xi.
\end{align*}
This complete the proof of theorem 1.

\section{The distribution discrepancies of $X_1$ and $X_4$}
\label{app3}
The two-sample Kolmogorov-Smirnov test for the distribution discrepancy under all selection mechanisms are significant (p $\approx$ 0).
\begin{figure}[H]
    \centering
    \includegraphics[width=0.8\textwidth]{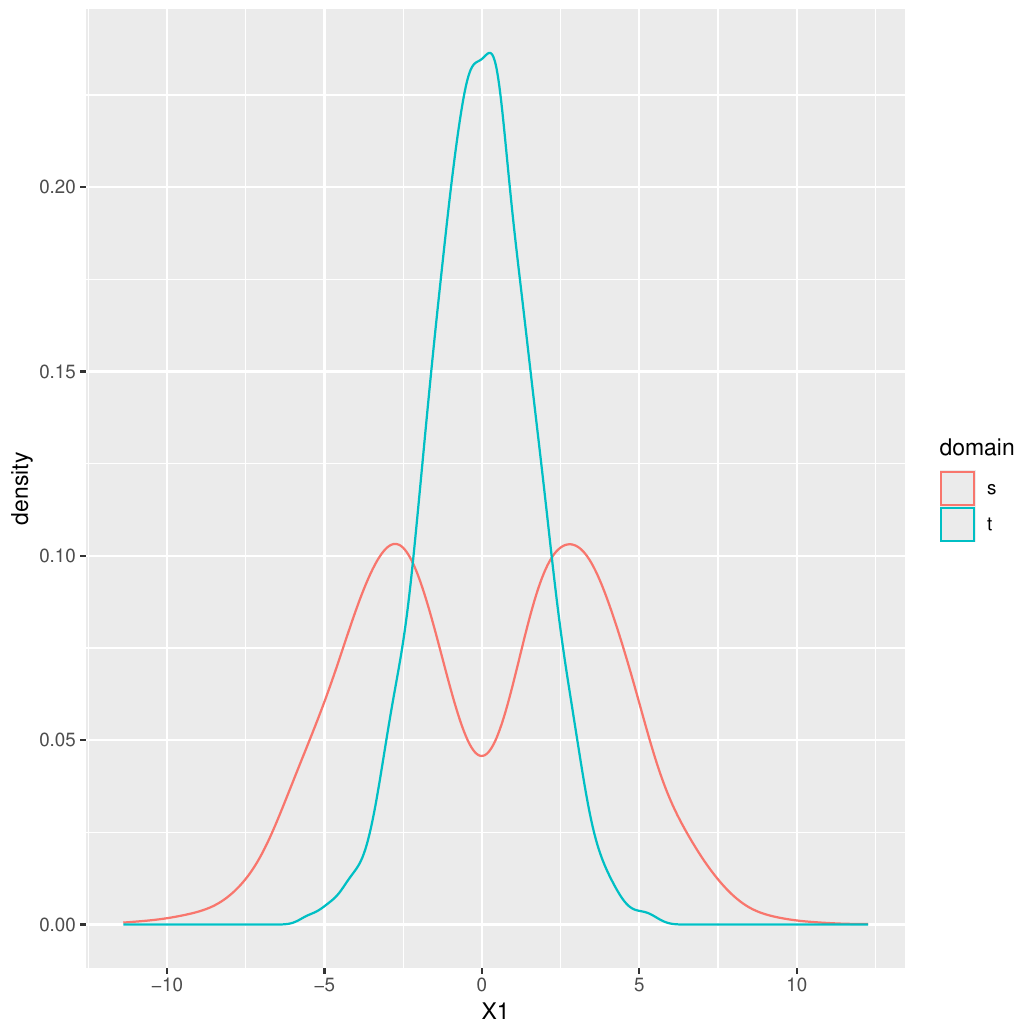}
    \caption{The distribution discrepancy of $X_1$ in source and target domains under the restricted range situation with $\texttt{score} = X_1$.}
    \label{fig6}
\end{figure}
\begin{figure}
 \centering
     \begin{subfigure}[b]{0.4\textwidth}
         \centering
         \includegraphics[width=\textwidth]{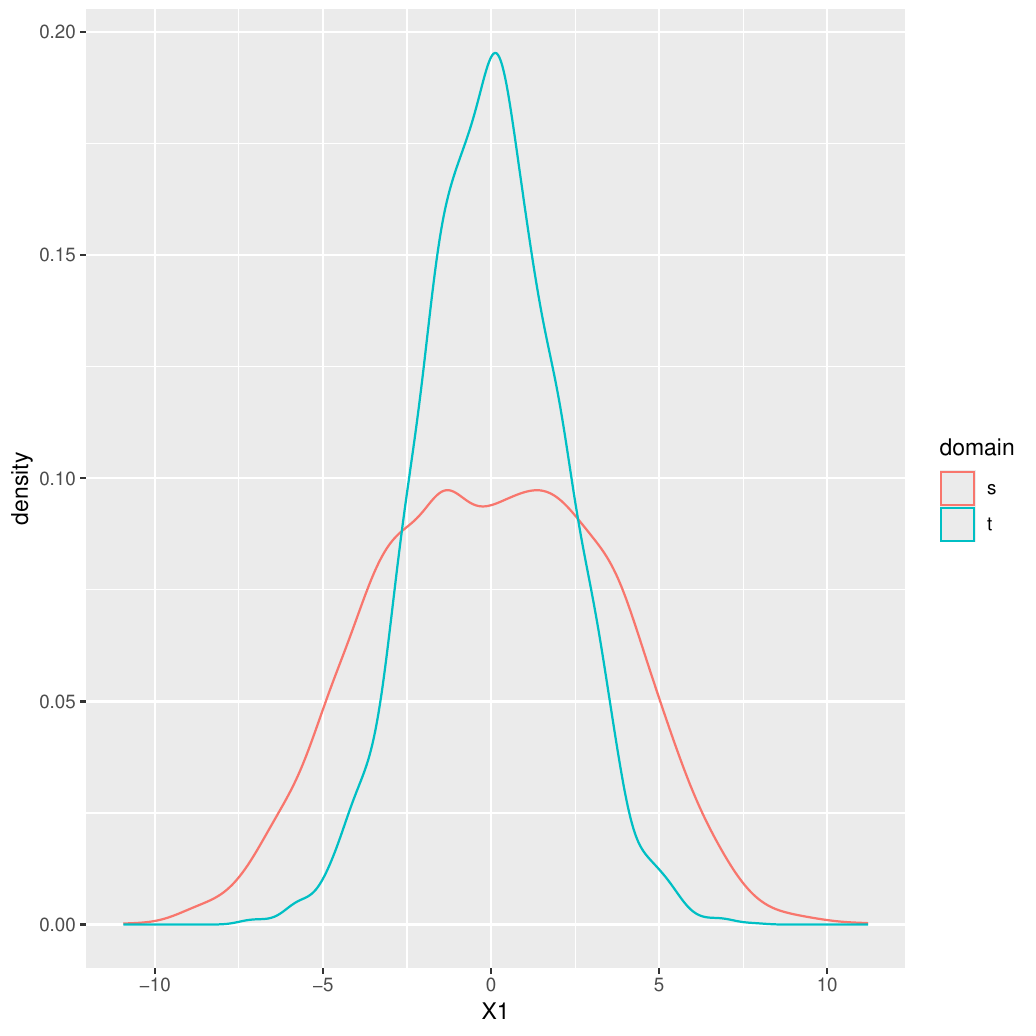}
         \caption{The distribution discrepancy of $X_1$.}
         \label{fig7a}
     \end{subfigure}
     \begin{subfigure}[b]{0.4\textwidth}
         \centering
         \includegraphics[width=\textwidth]{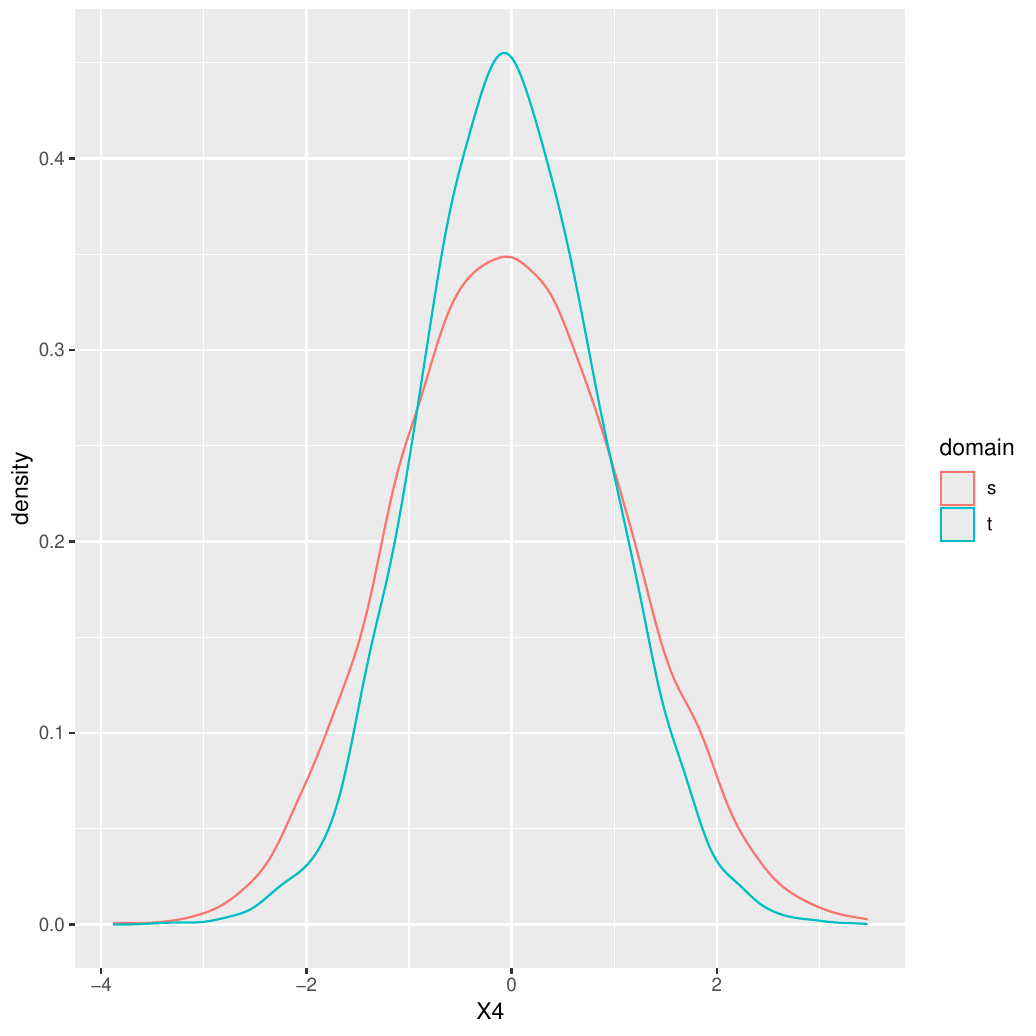}
         \caption{distribution discrepancy of $X_4$.}
         \label{fig7b}
     \end{subfigure}
     \caption{The distribution discrepancies of $X_1$ and $X_5$ in source and target domains under the restricted range situation with $\texttt{score} = X_1 + 2 * X_4$.}
        \label{fig7}
\end{figure}

\begin{figure}[H]
    \centering
    \includegraphics[width=0.8\textwidth]{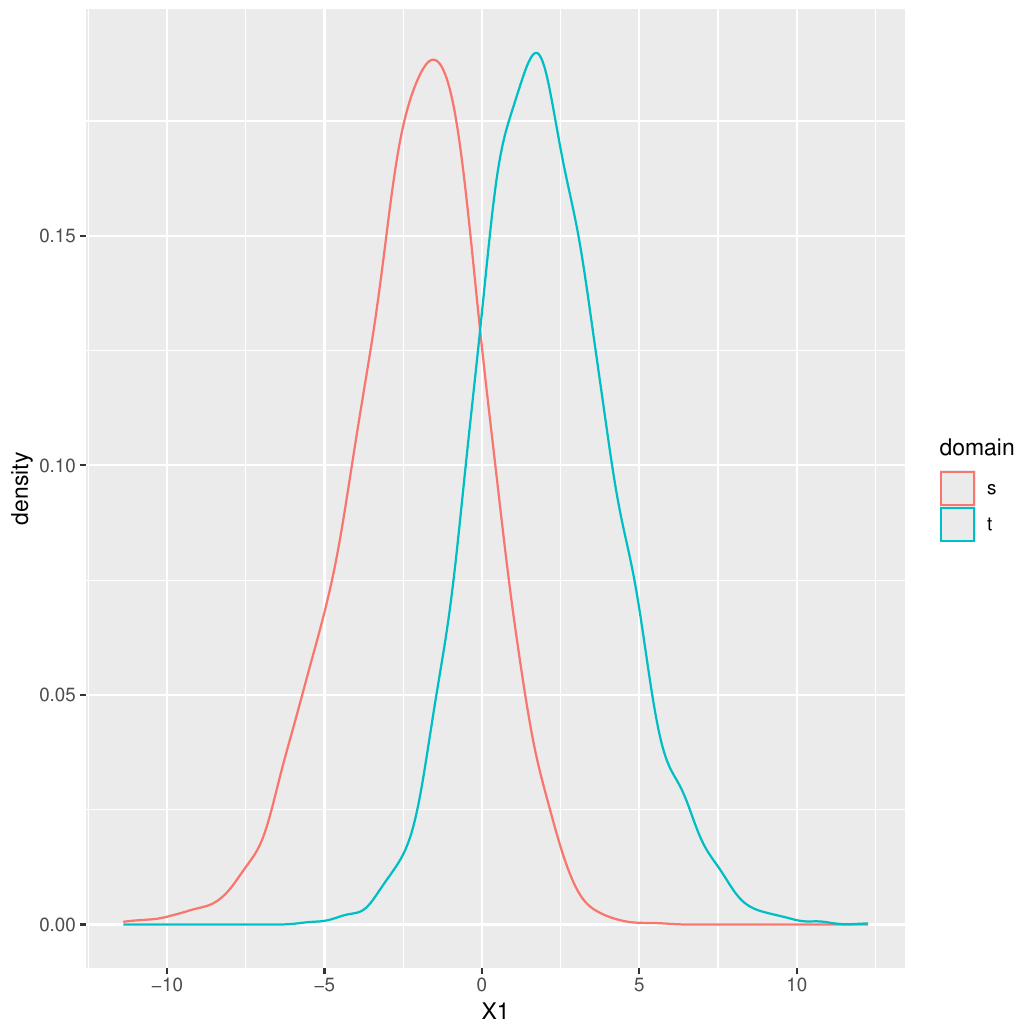}
    \caption{The distribution discrepancy of $X_1$ in source and target domains under the shifted location situation with $\texttt{score} = X_1$.}
    \label{fig8}
\end{figure}

\begin{figure}[H]
 \centering
     \begin{subfigure}[b]{0.4\textwidth}
         \centering
         \includegraphics[width=\textwidth]{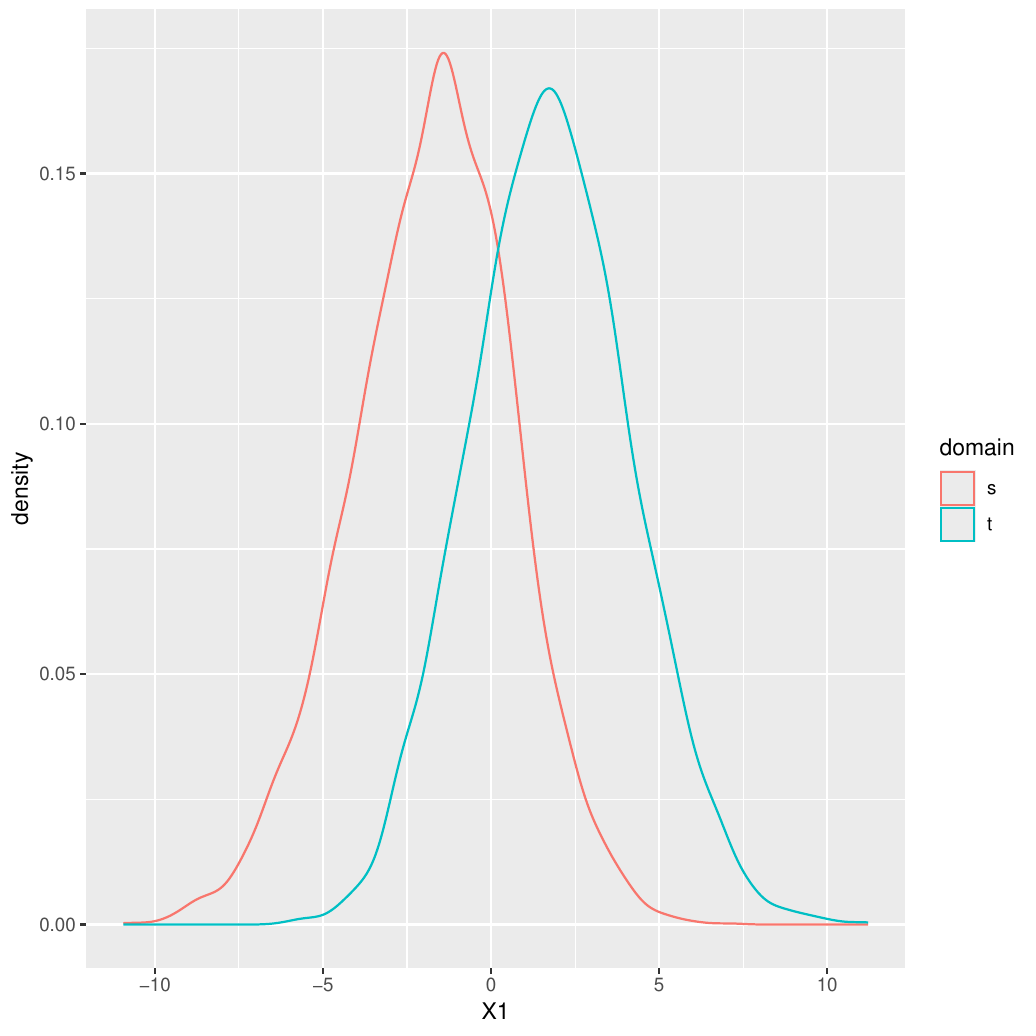 }
         \caption{The distribution discrepancy of $X_1$.}
         \label{fig9a}
     \end{subfigure}
     \begin{subfigure}[b]{0.4\textwidth}
         \centering
         \includegraphics[width=\textwidth]{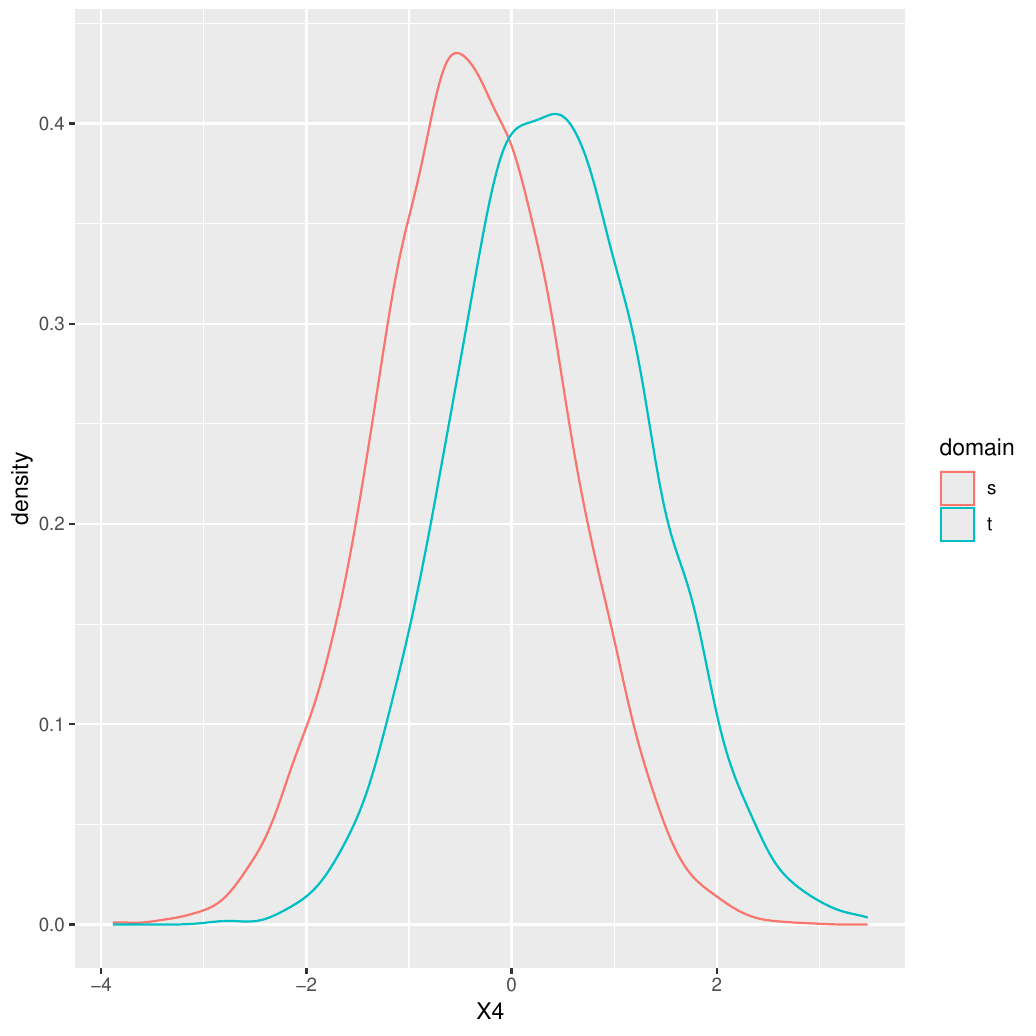}
         \caption{The distribution discrepancy of $X_4$.}
         \label{fig9b}
     \end{subfigure}
     \caption{The distribution discrepancies of $X_1$ and $X_5$ in source and target domains under the shifted location situation with $\texttt{score} = X_1 + 2 * X_4$.}
        \label{fig9}
\end{figure}

\section{Performance of DA-CART and DA-GLM-Tree Extensions}
\label{app4}
\begin{figure}[H]
    \centering
    \includegraphics[width=\textwidth]{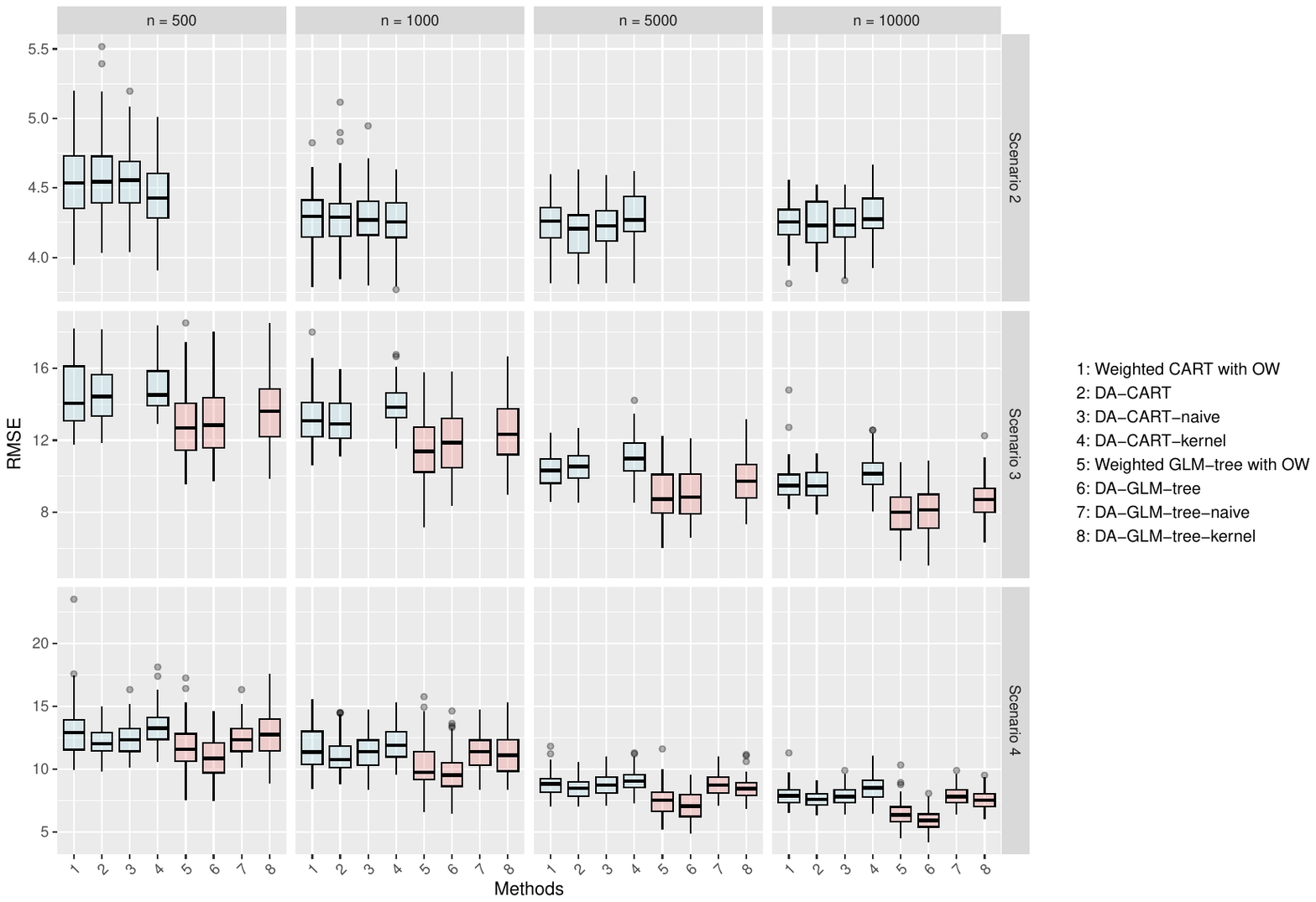}
    \caption{This figure presents boxplots of Root Mean Square Error (RMSE) across various tree models based on 100 repeated simulations. It consists of a 4 × 4 panel grid organized into columns representing different sample sizes (500, 1,000, 5,000, and 10,000) and rows for various scenarios. The scenarios are as follows: 2. Restricted range with sel $\subsetneq$ pred 3. Shifted range with sel $\subseteq$ pred 4. Shifted range with sel $\subsetneq$ pred. All methods being compared are labelled on the x-axis, and the corresponding model names are listed in the legend.}
\label{fig10}
\end{figure}
\section{Performance in non-discrepancy scenarios}
\label{app5}
\begin{figure}[H]
    \centering
    \includegraphics[width=\textwidth]{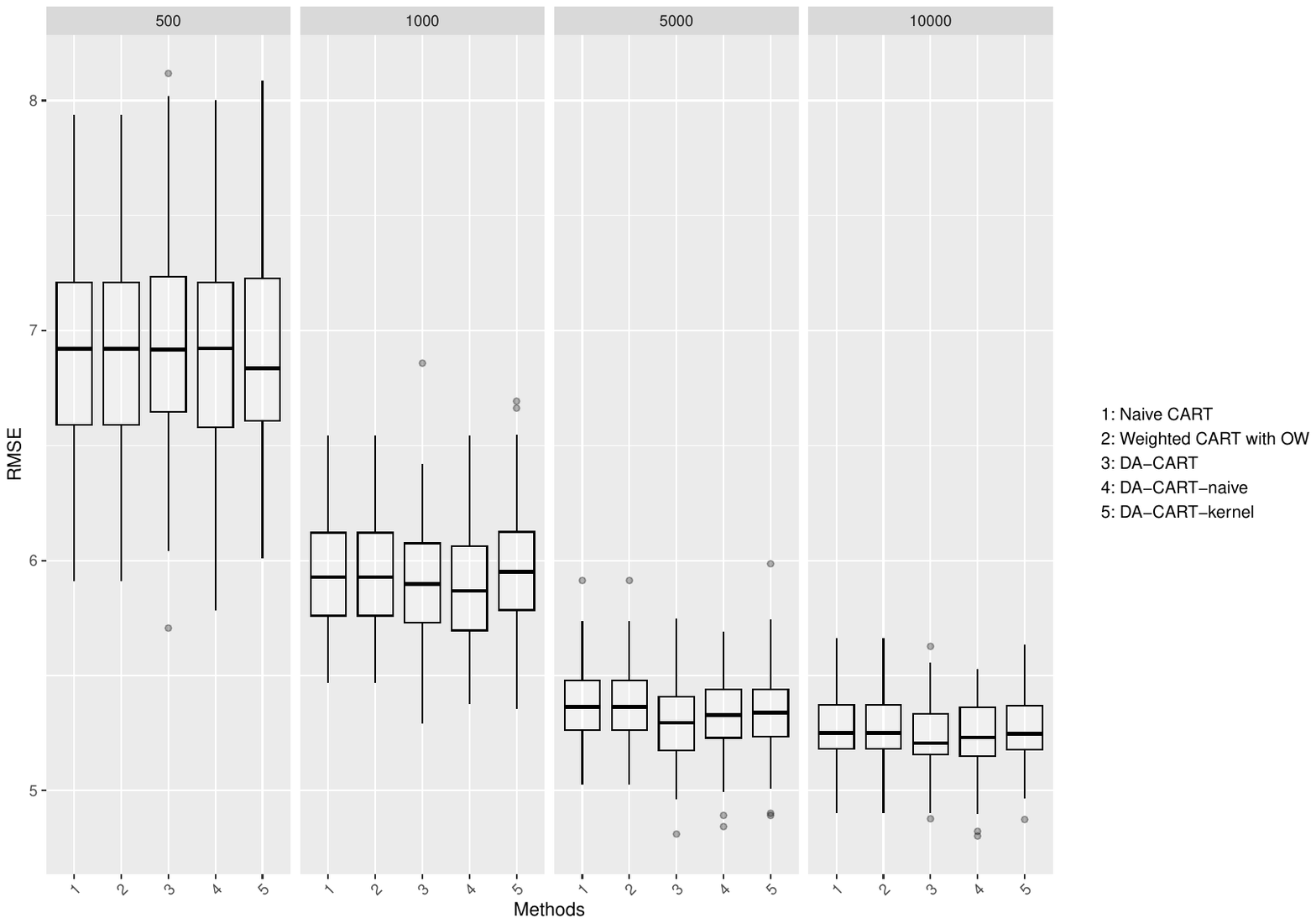}
    \caption{This figure displays boxplots of the Root Mean Square Error (RMSE) for various tree models based on 100 repeated simulations. It is organized into four columns, each corresponding to different sample sizes: 500, 1,000, 5,000, and 10,000. The methods being compared are labelled on the x-axis, and the names of the corresponding models are provided in the legend.}
    \label{fig11}
\end{figure}

\bibliographystyle{apacite}
\bibliography{transfer_learning.bib}

\end{document}